\documentclass{article}
\usepackage{ijcai25}
\usepackage{multirow}
\usepackage{times}
\usepackage{soul}
\usepackage{url}
\usepackage[hidelinks]{hyperref}
\usepackage[utf8]{inputenc}
\usepackage[ruled,vlined]{algorithm2e}
\usepackage[small]{caption}
\usepackage{graphicx}
\usepackage{amsmath}
\usepackage{amssymb}
\usepackage{amsthm}
\usepackage{booktabs}
\usepackage{algorithm2e}
\usepackage{enumitem}
\usepackage{algorithmic}
\usepackage[table]{xcolor} 
\definecolor{mygray}{gray}{0.92} 
\newcommand{\gc}{\cellcolor{mygray}}
\usepackage{adjustbox} % 修复 \adjustbox 错误
\usepackage[switch]{lineno}
\title{
LaPro-DTA: Latent Dual-View Drug Representations and Salient Protein Feature Extraction for Generalizable Drug--Target Affinity Prediction
}

\author{
    Zihan Dun$^1$\and
    Liuyi Xu$^1$\and
    An-Yang Lu$^1$\and
    Shuang Li$^2$\and
    Yining Qian$^{3*}$
    \affiliations
    $^1$College of Information Science and Engineering, Northeastern University, Shenyang 110819, China\\
    $^2$School of Artificial Intelligence, Beihang University, Beijing 100000, China\\
    $^3$School of Computer Science and Engineering, Northeastern University, Shenyang 110819, China
    \emails
    dunzihan@stumail.ysu.edu.cn,
    \{xuliuyi, luanyang\}@mails.neu.edu.cn,
    shuangliai@buaa.edu.cn,
    qianyiningning@126.com
}

\begin{document}
\maketitle
{\renewcommand{\thefootnote}{}\footnotetext{$^*$Corresponding author.}}

\begin{abstract}
Drug--target affinity prediction is pivotal for accelerating drug discovery, yet existing methods suffer from significant performance degradation in realistic cold-start scenarios (unseen drugs/targets/pairs), primarily driven by overfitting to training instances and information loss from irrelevant target sequences.
In this paper, we propose LaPro-DTA, a framework designed to achieve robust and generalizable DTA prediction.
To tackle overfitting, we devise a latent dual-view drug representation mechanism. 
It synergizes an instance-level view to capture fine-grained substructures with stochastic perturbation and a distribution-level view to distill generalized chemical scaffolds via semantic remapping, thereby enforcing the model to learn transferable structural rules rather than memorizing specific samples.
To mitigate information loss, we introduce a salient protein feature extraction strategy using pattern-aware top-$k$ pooling, which effectively filters background noise and isolates high-response bioactive regions. 
Furthermore, a cross-view multi-head attention mechanism fuses these purified features to model comprehensive interactions.
Extensive experiments on benchmark datasets demonstrate that LaPro-DTA significantly outperforms state-of-the-art methods, achieving an 8\% MSE reduction on the Davis dataset in the challenging unseen-drug setting, while offering interpretable insights into binding mechanisms.
\end{abstract}

\section{Introduction}
Drug--target affinity (DTA) measures the binding strength between drug molecules and target proteins, serving as a cornerstone in virtual screening and lead optimization~\cite{chen2025local,li2023structure}.
While traditional biochemical assays provide precise measurements, they are often prohibitively expensive and time-consuming, limiting their scalability for large-scale chemical space exploration~\cite{prasad2017research}.
Consequently, developing efficient data-driven computational methods has become indispensable for accelerating modern drug discovery.

% Drug--target affinity (DTA) measures the binding strength between a drug molecule and a target protein.
% Accurate DTA prediction is a key component of virtual screening and lead optimization, as it enables efficient prioritization of candidate compounds and significantly reduces reliance on costly wet-lab iterations~\cite{chen2025local,li2023structure}.
% Although traditional biochemical assays and high-throughput screening can provide precise affinity measurements, they are often prohibitively expensive and time-consuming, limiting their practicality for large-scale exploration~\cite{prasad2017research}.
% These bottlenecks highlight the need for efficient data-driven computational methods in DTA prediction.

% 为了提高DTA预测的准确性，近期研究希望通过引入重构任务的多任务学习方法或使用SMILES加分子图的多模态方法捕获更有效的药物特征。GDilatedDTA2024使用图膨胀卷积增强对分子结构信息的建模能力，PairVAE2025通过重建分子捕获分子结构信息，DMFF-DTA2025融合药物和靶标蛋白的序列与图结构信息提示亲和力预测性能。

The complexity of molecular interactions has driven a paradigm shift from classical machine learning to deep learning architectures~\cite{kim2021bayesian,watanabe2021deep}.
Early computational approaches mainly employed machine learning techniques for DTA prediction~\cite{ain2015machine,he2017simboost}.
With the adoption of deep learning, DTA prediction methods have increasingly moved toward end-to-end models that learn task-specific representations directly from raw input data \cite{yang2024interaction,li2020monn,zhao2020gansdta,fang2023colddta}.
DeepDTA~\cite{ozturk2018deepdta} used CNNs to model local patterns in protein and drug sequences, while GraphDTA~\cite{nguyen2021graphdta} adopted graph neural networks to represent molecular structures.
To improve the accuracy of DTA prediction, recent works have been proposed to investigate the capture of more effective drug features by incorporating multi-task learning methods with reconstruction tasks or using multi-model approaches~\cite{li2021co,zhang2024gdilateddta}.
For instance, \cite{13,shah2025deepdtagen} have been proposed to enhance feature representation via reconstruction, and DMFF-DTA~\cite{he2025dual} has been proposed to investigate the integration of sequence and graph structural information from both drugs and target proteins to improve affinity prediction performance.
Collectively, these methods have established strong baselines on standard benchmarks.

% 然而，大多数方法性能的提升主要是在随机划分下观察到的，在这种情况下，测试样本与训练样本中的药物和靶点存在重叠。
However, a critical gap remains:
their success is largely confined to random split settings, where test data follows the same distribution as training data.
% However, the performance improvements of most methods are primarily seen under random splits, where overlap exists between the drugs and targets in the test samples and the training set.
In realistic cold-start scenarios—where models must predict affinity for unseen drugs or targets—existing state-of-the-art methods suffer from significant performance degradation~\cite{huang2025gflearn,choppara2025q}.
% In practical drug discovery, models are often required to handle cold-start scenarios, where they must generalize to unseen data.
% This is particularly challenging, as models can nolonger rely on memorization or overfitting to training data, leading to a significant drop in performance~\cite{huang2025gflearn,choppara2025q}.
This generalization gap suggests that current methods often rely on memorizing training instances rather than learning the transferable chemical rules governing molecular interactions.
% This generalization gap isn't just a statistical issue. It reveals deeper problems in how to learn representations and how to retain important binding signals, making them more transferable.
We attribute this failure to two fundamental bottlenecks in current DTA representation learning:
\textbf{(1) Overfitting to instances:}
Existing approaches typically treat drug--target pairs as isolated labels. Lacking a mechanism to capture generalizable Structure-Activity Relationships (SAR)~\cite{bemis1996properties}, models tend to memorize specific training samples and their syntactic shortcuts rather than learning transferable chemical scaffolds. This overfitting is often exacerbated by auxiliary reconstruction tasks, which can introduce optimization conflicts distinct from the primary affinity objective~\cite{shah2025deepdtagen}.
% Current methods typically treat drug-target pairs as isolated instances, leading models to memorize specific training samples rather than learning generalized interaction rules. This results in representations tailored to seen data that fail to generalize to unseen entities~\cite{fang2023colddta}. The issue is further compounded by auxiliary reconstruction tasks, which introduce optimization conflicts that distract from the primary affinity objective~\cite{shah2025deepdtagen}. 
% These limitations severely hinder the model's transferability to cold-start scenarios.
\textbf{(2) Information loss from irrelevant target sequences:} 
Drug--target binding is inherently localized to specific active sites, whereas most methods compress the full-length target sequence into a single global representation. This mismatch causes salient binding signals to be diluted by the vast non-functional protein background~\cite{zhao2019attentiondta}. Without a mechanism to prioritize high-response features, models tend to average out critical local patterns with irrelevant protein segments, which severely compromises their capability to generalize to unseen targets~\cite{huang2021moltrans}.

To bridge this gap, we propose LaPro-DTA: \textbf{La}tent dual-view drug representations and salient \textbf{Pro}tein feature extraction, a novel framework designed for robust and generalizable DTA prediction.
% Firstly, we identify that the generalization gap in cold-start scenarios stems from two critical bottlenecks in representation learning: overfitting to instances and information loss.
We devise a dual-view representation learning mechanism for drugs and a salient feature extraction strategy for proteins, respectively.
Furthermore, a cross-view multi-head attention mechanism facilitates deep interaction between these purified representations for comprehensive affinity modeling.
% LaPro-DTA tackles the generalization challenge from three aligned angles: for drug, 
% mitigating overfitting via latent dual-view representation learning mechanism,
% for target, alleviating information loss through a salient feature extraction,
% and facilitating a comprehensive modeling via a cross-view multi-head attention fusion strategy.
% on the drug side, it reduces pair-specific memorization and improves global context transferability. 
% On the target side, it extracts pocket-related features to filter out the background, and then aligns dual-view representations to capture interaction evidence.
% Concretely, our framework \textbf{mitigates \underline{overfitting}} via a latent dual-view drug representation mechanism and \textbf{mitigates \underline{information loss}} through salient protein feature extraction, followed by cross-view alignment to integrate complementary features.
Our main contributions are:

\begin{itemize}[noitemsep, topsep=2pt, parsep=2pt]
    \item 
    We propose a latent dual-view drug representation framework that synergizes an instance-level view to capture fine-grained substructures and a distribution-level view to distill generalized chemical scaffolds via semantic remapping. By incorporating a stochastic encoding strategy, we provide a mathematically grounded solution acting as implicit regularization to effectively mitigate overfitting in data-scarce cold-start scenarios.
    % We introduce a latent dual-view drug representation framework that integrates an instance-level view and a distribution-level view.
    % Specifically, it combines the instance-level view focusing on specific details with the distribution-level view capturing general distributional features.
    % This architecture yields robust representations for cold-start prediction, especially for unseen drugs.
    % We introduce a stochastic instance-level detail latent with a distribution-level latent derived via remapping (encode $\rightarrow$ deconvolutional remap $\rightarrow$ re-encode).
    % This dual-view emphasizes transferable backbone patterns, yielding robust representations for cold-start prediction.
    
    \item 
    We develop a salient protein feature extraction strategy utilizing pattern-aware top-$k$ pooling.
    This mechanism functions as a semantic filter, explicitly discarding redundancy from non-binding regions while preserving the most discriminative bio-active features.
    Crucially, this selective retention mitigates the loss of critical information, facilitating precise interaction modeling.
    % Treating convolutional channels as pattern detectors, we selectively retain the Top-K activations per channel to suppress non-binding background while retaining high-confidence bioactive patterns, producing a purified key signal set for subsequent interaction modeling.

    \item 
    We conduct extensive evaluations on standard benchmarks and a strictly disjoint out-of-sample test set, demonstrating that LaPro-DTA establishes new state-of-the-art performance across unseen settings and offers superior interpretability through visualized residue-level saliency analysis.
    % We facilitate the deep interaction of the instance-level view and the distribution-level view with purified protein features by employing a cross-view multi-head attention mechanism.
    % This aggregates interaction evidence to ensure accurate affinity prediction.
    % We align the pattern-aware protein signals with each drug view separately and aggregate interaction evidence via a cross-view multi-head attention mechanism, enabling complementary use of instance chemical details and distribution representation for affinity prediction.

    % \item 
    % LaPro-DTA demonstrates superior generalization capabilities on cold-start datasets. Additionally, we provide intuitive visualization analysis to enhance the \textbf{interpretability} of the learned interaction logic.
\end{itemize}

\begin{figure*}[t]
    \centering
    \includegraphics[width=\textwidth, trim=5pt 5pt 5pt 5pt, clip]{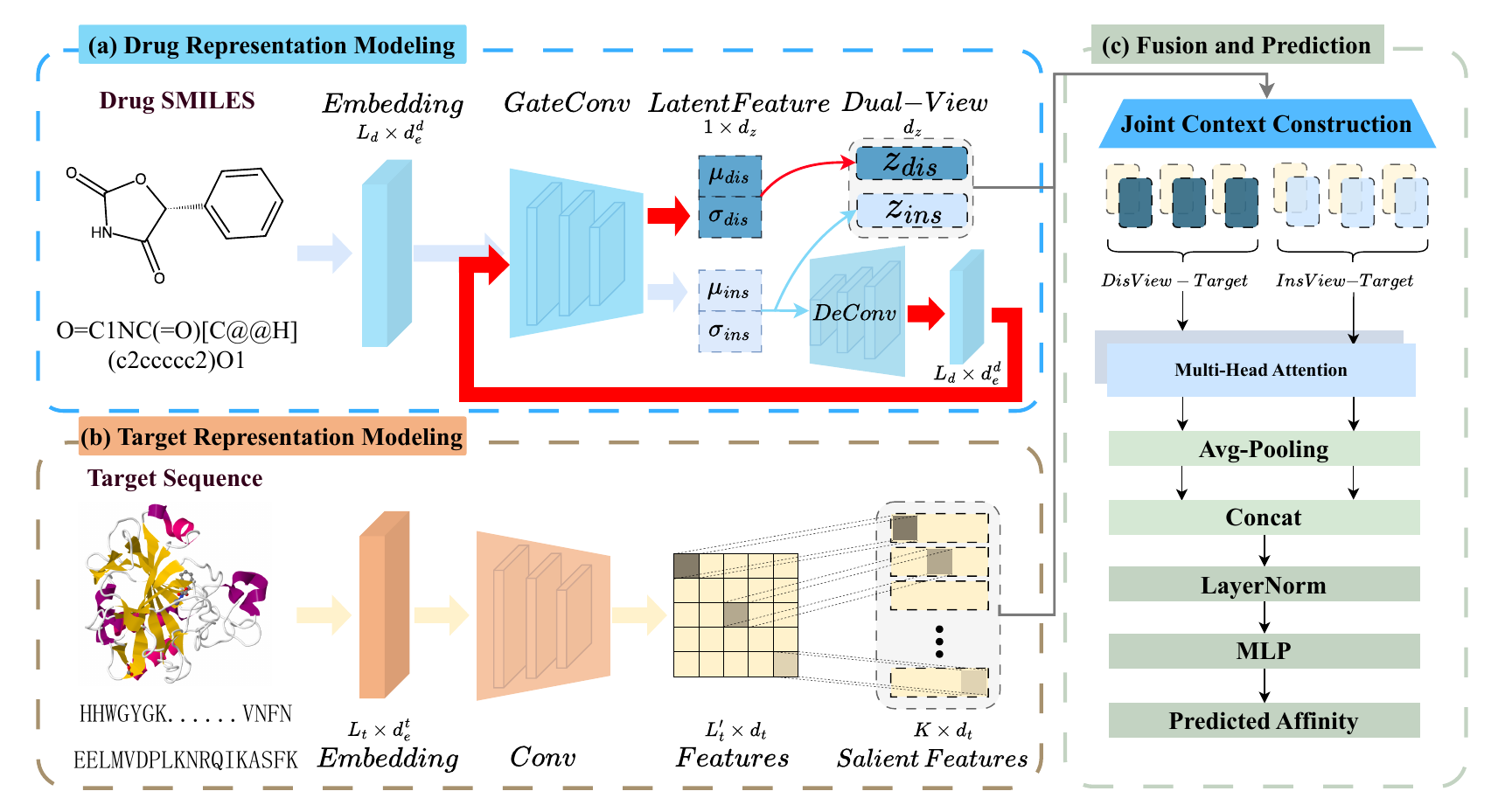}  % 左下右上 0.50 报too wide
    \caption
    {Overview of the LaPro-DTA architecture. (a) Drug Representation: Extracts drug features using a dual-view mechanism to capture both instance-level and distribution-level view. (b) Target Representation: Extracts target features by selectively focusing on key bioactive segments rather than the whole sequence. (c) Feature Fusion and Prediction: Aligns drug and target features to capture fine-grained interactions between them. The red line denotes the DeCNN-to-shared-encoder route that generates the distribution-level view. }
    \label{fig:overview}
\end{figure*}

\section{Methodology}
As illustrated in Figure~\ref{fig:overview}, LaPro-DTA comprises three core components: latent dual-view drug representation, salient protein feature extraction, and cross-view feature fusion with affinity prediction.

\subsection{Problem Formulation}
We formulate the DTA prediction task as a regression problem. 
Let $\mathcal{D} = \{d_1, d_2, \dots, d_N\}$ denote the set of drug molecules, where each drug is represented by its SMILES string \cite{weininger1988smiles}. 
Similarly, let $\mathcal{T} = \{t_1, t_2, \dots, t_M\}$ denote the set of target proteins, where each target is represented by its amino acid sequence. 
The interactions are provided as an affinity matrix $\mathbf{Y} \in \mathbb{R}^{N \times M}$, where $y_{ij}$ represents the continuous binding affinity value between drug $d_i$ and target $t_j$. 
The objective is to learn a predictive function $f: \mathcal{D} \times \mathcal{T} \rightarrow \mathbb{R}$ that accurately estimates the affinity score for a given drug-target pair. 
Crucially, under cold-start settings, the drug $d$ or target $t$ encountered at inference time is unseen during training.
\subsection{Drug Representation Modeling and Latent Dual-View Mechanism}

\noindent \textbf{Main Idea.} 
To mitigate \textbf{overfitting}, we introduce a latent dual-view mechanism. 
This design discourages the model from capturing syntactic shortcuts, shifting its focus to features governed by distributional regularities.

\noindent \textbf{Instance-Level View.} 
We define a shared encoder $\mathcal{E}$ to extract latent representations of drugs. Specifically, $\mathcal{E}$ first employs a GatedCNN block \cite{dauphin2017language} to capture local substructural features from the input drug embedding $\mathbf{X}_d \in \mathbb{R}^{L_d \times d^{d}_e}$. This block consists of three stacked convolutional layers where information flow is controlled by the element-wise product of a linear convolutional feature map and a sigmoid-activated gate. Subsequently, the encoder projects these features into latent distribution parameters: the mean $\boldsymbol{\mu}_{ins} \in \mathbb{R}^{d_z}$ and the log-variance $\mathbf{h}_{var} \in \mathbb{R}^{d_z}$.

To enhance representation robustness, we propose a \textbf{stochastic encoding strategy (SES)}. The instance-level sampling process is defined as:
\begin{equation}
    \mathbf{z}_{ins} = \boldsymbol{\mu}_{ins} + \boldsymbol{\epsilon} \odot \boldsymbol{\sigma}_{ins}, \quad \boldsymbol{\epsilon} \sim \mathcal{N}(\mathbf{0}, \mathbf{I})
\end{equation}
where $\boldsymbol{\sigma}_{ins} = \lambda \cdot \exp\left(\frac{1}{2}\text{ReLU}(\mathbf{h}_{var})\right)$. Here, we enforce a hard architectural constraint using a ReLU activation to ensure the standard deviation has a lower bound $\lambda$, preventing the latent space from collapsing into deterministic points.
Theoretically, this stochastic sampling under the lower-bound constraint acts as an implicit Tikhonov regularization \cite{Bishop1995}. To demonstrate this, let $F: \mathbb{R}^{d_z} \to \mathbb{R}$ denote the downstream affinity prediction function. To isolate the regularization effect, we temporarily focus on the instance-level view $\mathbf{z}_{ins}$ by treating other views and target features as constants. By applying a first-order Taylor expansion to $F$ around $\boldsymbol{\mu}_{ins}$, the expected loss under perturbation $\boldsymbol{\epsilon}$ can be approximated as:
\begin{equation}
\begin{aligned}
\mathbb{E}_{\boldsymbol{\epsilon}}[\tilde{\mathcal{L}}] 
&\approx \mathbb{E}_{\boldsymbol{\epsilon}}\left[ (y - F(\boldsymbol{\mu}_{ins}) - \nabla F^\top (\boldsymbol{\sigma}_{ins} \odot \boldsymbol{\epsilon}))^2 \right] \\
&= \mathcal{L}_{MSE} + \sum\nolimits_i (\sigma_{ins,i} \cdot \partial_{z_i} F)^2 \\
&\ge \mathcal{L}_{MSE} + \lambda^2 \|\nabla F(\boldsymbol{\mu}_{ins})\|_2^2
\end{aligned}
\end{equation}
where the cross-terms vanish since $\mathbb{E}[\boldsymbol{\epsilon}]=\mathbf{0}$. The constraint $\boldsymbol{\sigma}_{ins} \ge \lambda$ imposes a strict penalty on the gradient norm $\|\nabla F\|_2^2$. This effectively smoothes the decision boundaries of the prediction function, thereby preventing overfitting in data-scarce cold-start settings.

\noindent \textbf{Distribution-Level View.} 
Relying solely on instance-level representations suffers from a critical limitation: models tend to memorize specific input details, failing to learn the generalized regularities necessary for extrapolating to unseen drug scenarios. To bridge this gap, we introduce a distribution-level view designed to distill generalized structural regularities from the chemical space.
Formally, we employ a DeCNN module~\cite{zeiler2010deconvolutional} to map the compressed instance latent $\mathbf{z}_{ins}$ to an intermediate feature map $\mathbf{H}_{remap} \in \mathbb{R}^{L_d \times d^{d}_e}$. Distinct from autoencoders that depend on token-wise reconstruction losses to memorize inputs, our approach performs a \textit{task-driven semantic remapping}. We feed $\mathbf{H}_{remap}$ back into the shared encoder $\mathcal{E}$ to generate the final distribution-level representation $\mathbf{z}_{dis}$:
\begin{equation}
    [\boldsymbol{\mu}_{dis}, \boldsymbol{\sigma}_{dis}] = \mathcal{E}(\mathbf{H}_{remap})
\end{equation}
\begin{equation}
    \mathbf{z}_{dis} = \boldsymbol{\mu}_{dis} + \boldsymbol{\epsilon} \odot \boldsymbol{\sigma}_{dis}, \quad \boldsymbol{\epsilon} \sim \mathcal{N}(0, \mathbf{I})
\end{equation}
In this process, by eliminating the constraint of token-wise reconstruction, we decouple feature learning from the memorization of trivial input details. Instead, under the exclusive guidance of the discriminative DTA objective, the remapping process is compelled to distill only those distributional regularities that are predictively critical for drug-target interactions. This mechanism effectively filters out instance-specific noise while retaining the robust chemical scaffolds necessary for accurate affinity modeling, thereby providing a distribution-level perspective with enhanced generalization capabilities.

\noindent \textbf{Dual-View Synergy.} 
Guided by the SAR principle~\cite{bemis1996properties}, our mechanism addresses the challenge where similar structures share bioactivities yet subtle variations alter affinity. The distribution-level view captures generalized chemical scaffolds, enabling inference on unseen entities via broad structural regularities. Complementarily, the instance-level view resolves fine-grained atomic details, distinguishing specific interactions within local substructures. Integrating these perspectives balances the extrapolation capability needed for cold-start scenarios with the discrimination precision required for accurate prediction.
\subsection{Salient Target Features Extraction with Pattern-Aware Top-K Pooling}

\textbf{Main Idea.} 
To mitigate \textbf{information loss} caused by global compression, we propose a salient feature extraction strategy that acts as a selective filter.
Using top-$k$ pooling, this mechanism prioritizes high-activation features, explicitly retaining only the most discriminative local patterns while filtering out redundancy.
We use stacked CNNs to extract local representations from the target protein embedding $\mathbf{X}_t \in \mathbb{R}^{L_t \times d_e}$. 
With this architecture, the convolutional filters act as specific feature detectors, capturing contextualized local structures (\textit{e.g.}, motifs) into feature channels.

\noindent \textbf{Pattern-Aware Top-$K$ Pooling.} \cite{kalchbrenner2014convolutional} demonstrated the effectiveness of $k$-max pooling in capturing the most informative sequence elements.
Drawing on this, we adapt this mechanism to construct compact \textbf{salient features}. 
To retain high-value information, we adopt a high-activation priority strategy. 
This design explicitly preserves dominant residue sets, ensuring that the downstream attention mechanism focuses on the most reactive biochemical patterns.
Formally, let $\mathbf{H}_{conv} \in \mathbb{R}^{L^{\prime}_t\times d_t }$ denote the convolutional feature map, where $d_t$ is the number of filters. 
We identify the $K$ most significant activations for each channel independently:
\begin{equation} \label{eq:topk_pooling}
    \mathbf{H}_t = \mathrm{TopK}(\mathbf{H}_{conv}, K)
\end{equation}
% The resulting feature map $\mathbf{H}_{raw} \in \mathbb{R}^{d_h \times K}$ is then transposed to $\mathbf{H}_t \in \mathbb{R}^{K \times d_h}$. 
This representation $\mathbf{H}_t \in \mathbb{R}^{K \times d_t}$ aggregates  the top-$k$ activations, where each entry corresponds to a critical residue within the sequence.
This hyperparameter selection aligns with the sparsity assumption of binding pockets, suggesting that identifying the top few strongest activations per channel is sufficient to characterize the presence of specific biochemical motifs while suppressing redundant information.
Finally, the salient features serve as the input for the fusion module. This ranking-based profile aggregates the strongest responses across all pattern detectors, robustly capturing key binding evidence. This capability is corroborated by our analysis of residue-level saliency maps in Section 3.8.

\subsection{Cross-View Multi-Head Attention Fusion and Affinity Prediction}

\textbf{Main Idea.}
To leverage the dual-view drug representations and high-response protein features, we propose a cross-view multi-head attention fusion module. 
Unlike global pooling which causes attention degeneration, our top-$k$ strategy preserves salient protein features, providing granular context for fine-grained interaction modeling.

\noindent \textbf{Joint Context Construction.}
Drug vectors ($\mathbf{z}_{ins}$ and $\mathbf{z}_{dis}$) are broadcasted to match the cardinality $K$ of the protein segments ($\mathbf{H}_t$) and concatenated, followed by a linear projection to form context matrices $\mathbf{C}_{ins}\in \mathbb{R}^{K \times d_t}$ and $\mathbf{C}_{dis}\in \mathbb{R}^{K \times d_t}$. This design addresses two realistic scenarios: (1) instance-level detail context ($\mathbf{C}_{ins}$) captures precise atom-level details for identifying trained or similar drugs; (2) distribution-level representation context ($\mathbf{C}_{dis}$) facilitates cold-start generalization by leveraging scaffold similarity to infer binding potential for unseen molecules.
The drug representations $\mathbf{z}_{ins}, \mathbf{z}_{dis} \in \mathbb{R}^{d_z}$ are spatially aligned with the protein segments $\mathbf{H}_t \in \mathbb{R}^{K \times d_t}$ via broadcasting. The construction of the local context matrix $\mathbf{C}_{ins}$ is defined as:
\begin{equation}
    \mathbf{C}_{ins} = \phi \left( [ \mathbf{1}_K \mathbf{z}_{ins}^\top \parallel \mathbf{H}_t ] \mathbf{W}_i + \mathbf{b}_i \right)
\end{equation}
where $\mathbf{1}_K \in \mathbb{R}^{K \times 1}$ is a vector of ones, and $\parallel$ denotes the concatenation along the feature dimension. Similarly, the distribution-level representation matrix $\mathbf{C}_{dis}$ is formulated as:
\begin{equation}
    \mathbf{C}_{dis} = \phi \left( [ \mathbf{1}_K \mathbf{z}_{dis}^\top \parallel \mathbf{H}_t ] \mathbf{W}_d + \mathbf{b}_d \right)
\end{equation}
In both cases, $\mathbf{W}_i, \mathbf{W}_d \in \mathbb{R}^{(d_z + d_t) \times d_t}$ are learnable projection matrices and $\phi(\cdot)$ is the non-linear activation function.
\noindent \textbf{Attention Fusion and Prediction.}
We employ multi-head attention where protein segments $\mathbf{H}_t$ serve as the Query ($\mathbf{Q}$), while the joint contexts act as both Key ($\mathbf{K}$) and Value ($\mathbf{V}$). For the instance-level view interaction:
\begin{equation}
    \mathbf{Q} = \mathbf{H}_t, \quad \mathbf{K} = \mathbf{V} = \mathbf{C}_{ins}, \quad \mathbf{O}_{ins} = \mathrm{Softmax}\left(\frac{\mathbf{Q} \mathbf{K}^\top}{\sqrt{d_t}}\right) \mathbf{V}
\end{equation}
The global output $\mathbf{O}_{dis}$ is computed analogously. Finally, the affinity $\hat{y}$ is predicted by aggregating the dual-view features:
\begin{equation}
    \hat{y} = \mathrm{MLP}\left( \mathrm{LN}\left( \left[ \mathrm{Mean}(\mathbf{O}_{ins}) \mathbin{\|} \mathrm{Mean}(\mathbf{O}_{dis}) \right] \right) \right)
\end{equation}

\noindent \textbf{Optimization Objective:}
We adopt the Mean Squared Error (MSE) as the objective function to minimize the discrepancy between the ground-truth affinity $y_i$ and the predicted value $\hat{y}_i$. The loss $\mathcal{L}$ is defined as:
\begin{equation}
    \mathcal{L}_{MSE} = \frac{1}{N} \sum_{i=1}^{N} (y_i - \hat{y}_i)^2
\end{equation}

\begin{table}[ht]
    \centering
    \small 
    \begin{tabular}{lllll}
        \toprule
        Dataset & Drugs & Targets & Interactions & Affinity metrics \\
        \midrule
        KIBA      & 2,111    & 229   & 118,254 & $K_d,K_i\,and\,IC_{50}$  \\
        % BindingDB & 17,980   & 1,839 & 56,527  \\
        Davis     & 68       & 442   & 30,056 & $K_d$  \\
        \bottomrule
    \end{tabular}
    \caption{Summary of the datasets used in our experiments.}
    \label{tab:datasets}
    
\end{table}

\begin{table*}[!ht]
    \centering
    \scriptsize % 保持小号字体
    \renewcommand{\arraystretch}{0.9} %稍微增加一点行高以适应变宽的列
    \setlength{\tabcolsep}{5pt} % 增加列间距，使表格更宽

    \begin{tabular}{lcccccccc}
        \toprule
        \multirow{2}{*}{\textbf{Model}} & \multicolumn{4}{c}{\textbf{Davis}} & \multicolumn{4}{c}{\textbf{KIBA}} \\
        \cmidrule(lr){2-5} \cmidrule(lr){6-9}
        & \textbf{MSE (std)} $\downarrow$ & \textbf{MAE (std)} $\downarrow$ & \textbf{CI (std)} $\uparrow$ & $\mathbf{r_m^2}$ \textbf{(std)} $\uparrow$ & \textbf{MSE (std)} $\downarrow$ & \textbf{MAE (std)} $\downarrow$ & \textbf{CI (std)} $\uparrow$ & $\mathbf{r_m^2}$ \textbf{(std)} $\uparrow$ \\
        \midrule
        
        % =================== Setting: Unseen Drug ===================
        \multicolumn{9}{c}{\cellcolor{gray!10}\textit{\textbf{Setting: Unseen Drug}}} \\ 
        \midrule
        DeepDTA (2018) & 0.745 (0.008) & 0.704 (0.050) & 0.699 (0.045) & 0.078 (0.023) & 0.478 (0.007) & 0.485 (0.004) & 0.688 (0.006) & 0.263 (0.002) \\
        AttentionDTA (2019) & 0.656 (0.004) & 0.622 (0.003) & 0.682 (0.010) & 0.104 (0.004) & 0.433 (0.041) & 0.432 (0.031) & 0.722 (0.025) & 0.322 (0.030) \\
        Co-VAE (2021) & 0.577 (0.009) & 0.523 (0.001) & 0.616 (0.005) & 0.128 (0.014) & 0.429 (0.019) & 0.459 (0.041) & \underline{0.745 (0.010)} & \underline{0.334 (0.019)} \\
        GraphDTA (2021) & 0.590 (0.016) & 0.552 (0.014) & 0.695 (0.008) & 0.077 (0.010) & 0.473 (0.021) & 0.526 (0.020) & 0.738 (0.004) & 0.282 (0.032) \\
        ColdDTA (2023) & 0.613 (0.012) & 0.522 (0.008) & 0.716 (0.020) & \underline{0.146} (0.012) & \underline{0.427 (0.018)} & \underline{0.448 (0.030)} & 0.722 (0.008) & 0.329 (0.029) \\
        TransVAE-DTA (2024) & 0.654 (0.030) & 0.559 (0.016) & 0.570 (0.021) & 0.030 (0.017) & 0.537 (0.047) & 0.501 (0.036) & 0.715 (0.024) & 0.201 (0.031) \\
        % MF-DTA (2025) & \underline{0.573 (0.046)} & -- & 0.696 (0.017) & \underline{0.116 (0.013)} & -- & -- & -- & -- \\
        % MultiKD-DTA (2025) & 0.586 (0.000) & 0.606 (0.000) & 0.635 (0.000) & 0.084 (0.000) & -- & -- & -- & -- \\
        PairVAE (2025) & \underline{0.566 (0.004)} & \underline{0.498 (0.007)} & \underline{0.730 (0.003)} & 0.134 (0.036) & 0.469 (0.031) & 0.469 (0.016) & 0.727 (0.007) & 0.328 (0.024) \\
        \gc\textbf{LaPro-DTA (ours)} & \gc\textbf{0.519 (0.010)} & \gc\textbf{0.454 (0.009)} & \gc\textbf{0.737 (0.001)} & \gc\textbf{0.191 (0.028)} & \gc\textbf{0.403 (0.014)} & \gc\textbf{0.409 (0.014)} & \gc\textbf{0.755 (0.004)} & \gc\textbf{0.352 (0.024)} \\

        % =================== Setting: Unseen Target ===================
        \midrule
        \multicolumn{9}{c}{\cellcolor{gray!10}\textit{\textbf{Setting: Unseen Target}}} \\ 
        \midrule
        DeepDTA (2018) & 0.884 (0.024) & 0.711 (0.008) & 0.751 (0.033) & 0.183 (0.015) & 0.542 (0.012) & 0.498 (0.019) & 0.706 (0.015) & 0.365 (0.024) \\
        AttentionDTA (2019) & 0.565 (0.014) & 0.560 (0.022) & 0.814 (0.002) & 0.450 (0.033) & 0.543 (0.029) & 0.493 (0.013) & 0.704 (0.021) & 0.349 (0.023) \\
        Co-VAE (2021) & 0.570 (0.018) & 0.518 (0.016) & 0.811 (0.011) & \underline{0.443 (0.010)} & 0.531 (0.013) & 0.523 (0.020) & 0.712 (0.005) & 0.351 (0.015) \\
        GraphDTA (2021) & 0.565 (0.010) & 0.527 (0.011) & 0.763 (0.003) & 0.399 (0.014) & 0.501 (0.030) & 0.531 (0.026) & 0.665 (0.014) & 0.241 (0.040) \\
        ColdDTA (2023) & \underline{0.554 (0.008)} & \underline{0.457 (0.012)} & \underline{0.818 (0.024)} & 0.428 (0.020) & \underline{0.491 (0.018)} & 0.515 (0.030) & 0.685 (0.027) & 0.259 (0.012) \\
        TransVAE-DTA (2024)& 0.834 (0.034) & 0.559 (0.012) & 0.728 (0.017) & 0.201 (0.022) & 0.500 (0.038) & 0.524 (0.047) & 0.651 (0.038) & 0.208 (0.022) \\
        % MF-DTA (2025) & \underline{0.566 (0.004)} & \underline{0.498 (0.007)} & \underline{0.730 (0.003)} & \underline{0.134 (0.036)} & -- & -- & -- & -- \\
        % MultiKD-DTA (2025) & 0.608 (0.000) & 0.564 (0.000) & 0.816 (0.000) & 0.442 (0.000) & -- & -- & -- & -- \\
        PairVAE (2025)& 0.637 (0.010) & 0.497 (0.009) & 0.813 (0.006) & 0.373 (0.050) & 0.516 (0.033) & \underline{0.502 (0.005)} & \underline{0.716 (0.014)} & \underline{0.371 (0.025)} \\
        \gc\textbf{LaPro-DTA (ours)} & \gc\textbf{0.519 (0.016)} & \gc\textbf{0.449 (0.007)} & \gc\textbf{0.828 (0.007)} & \gc\textbf{0.473 (0.021)} & \gc\textbf{0.489 (0.010)} & \gc\textbf{0.491 (0.015)} & \gc\textbf{0.726 (0.006)} & \gc\textbf{0.403 (0.019)} \\

        % =================== Setting: Unseen Pair ===================
        \midrule
        \multicolumn{9}{c}{\cellcolor{gray!10}\textit{\textbf{Setting: Unseen Pair}}} \\ 
        \midrule
        DeepDTA (2018)& 0.886 (0.027) & 0.737 (0.012) & 0.712 (0.030) & 0.072 (0.030) & 0.708 (0.035) & 0.590 (0.004) & 0.631 (0.009) & 0.166 (0.020) \\
        AttentionDTA (2019)& 0.837 (0.020) & 0.696 (0.002) & 0.715 (0.006) & \underline{0.114 (0.006)} & 0.674 (0.036) & 0.564 (0.018) & 0.631 (0.018) & 0.165 (0.027) \\
        Co-VAE (2021)& 0.828 (0.026) & 0.617 (0.025) & 0.597 (0.011) & 0.096 (0.005) & 0.640 (0.005) & 0.565 (0.019) & 0.658 (0.006) & 0.185 (0.002) \\
        GraphDTA (2021)& 0.874 (0.036) & 0.606 (0.023) & 0.631 (0.021) & 0.032 (0.020) & 0.643 (0.033) & 0.573 (0.016) & 0.597 (0.013) & 0.055 (0.008) \\
        ColdDTA (2023) & \underline{0.819 (0.012)} & 0.622 (0.008) & 0.654 (0.014) & 0.117 (0.013) & 0.629 (0.010) & 0.570 (0.016) & 0.641 (0.008) & 0.170 (0.014) \\
        TransVAE-DTA (2024)& 0.911 (0.039) & 0.632 (0.013) & 0.554 (0.020) & 0.018 (0.012) & 0.692 (0.018) & 0.590 (0.038) & 0.497 (0.003) & 0.071 (0.033) \\
        % DGCA-DTA (2025) & \underline{0.566 (0.004)} & \underline{0.498 (0.007)} & \underline{0.730 (0.003)} & \underline{0.134 (0.036)} & -- & -- & -- & -- \\
        % MultiKD-DTA (2025) & 0.990 (0.000) & 0.697 (0.000) & 0.616 (0.000) & 0.022 (0.000) & -- & -- & -- & -- \\
        PairVAE (2025)& 0.820 (0.032) & \underline{0.582 (0.001)} & \underline{0.741 (0.010)} & 0.010 (0.003) & \textbf{0.621 (0.020)} & \underline{0.560 (0.006)} & \textbf{0.672 (0.009)} & \textbf{0.201 (0.012)} \\
        \gc\textbf{LaPro-DTA (ours)} & \gc\textbf{0.727 (0.016)} & \gc\textbf{0.546 (0.006)} & \gc\textbf{0.770 (0.019)} & \gc\textbf{0.166 (0.020)} & \gc\underline{0.637 (0.018)} & \gc\textbf{0.558 (0.014)} & \gc\underline{0.660 (0.005)} & \gc\underline{0.194 (0.016)} \\
        
        \bottomrule
    \end{tabular}
    \caption{Performance comparison on Davis and KIBA datasets under cold-start scenarios. (\textbf{Best}, \underline{Second Best}). Results are reported as mean (standard deviation) over five independent runs.}
    \label{tab:coldstart}
    
\end{table*}

\section{Experiments and Results}

\subsection{Datasets}

We evaluate LaPro-DTA on two benchmark datasets: Davis~\cite{davis2011comprehensive} and KIBA~\cite{tang2014making}. Following standard protocols, the interaction values in Davis are converted into their negative logarithmic forms ($pK_d$). Detailed statistics are summarized in Table~\ref{tab:datasets}.

\noindent \textbf{Settings.} 
To rigorously assess generalization, we design two distinct experimental protocols:

\noindent \textbf{Cold-Start Setting (Primary Focus).}
To simulate realistic drug discovery, we randomly reserve 20\% of drugs and targets as unseen entities. Based on this partition, the dataset is systematically stratified into three distinct scenarios: unseen drug (test drugs $\times$ training targets), unseen target (training drugs $\times$ test targets), and unseen pair (both unseen). For each scenario, the interactions are split equally into validation and testing sets, ensuring that test entities remain strictly unobserved during training.

\noindent \textbf{Random Split Setting.}
Additionally, we conduct experiments under the standard random split setting (randomly dividing interactions into train/validation/test sets) to serve as a performance reference. Detailed results for this setting are provided in Table S2 of the Supplementary Material.
% We also observe consistent improvements under the standard Random Split setting following [Zhang et al., 2024]. Davis and KIBA are divided into training and
% test sets with a 5:1 ratio, while BindingDB uses an 8:2 ratio.

\subsection{Experimental Details}
\noindent \textbf{Hyperparameter Settings.}
LaPro-DTA is implemented in PyTorch 2.0 and trained on a single NVIDIA RTX 3090 GPU. 
The model is optimized using Adam with a learning rate of $5 \times 10^{-4}$ and a batch size of 256. 
To mitigate overfitting, we employ an early stopping strategy with a patience of 20 epochs (maximum 100 epochs). 
The Drug Encoder employs a 3-block Gated CNN (kernel size 4) with a latent dimension of $d=96$. The Protein Encoder consists of a 3-layer CNN followed by top-$k$ pooling ($k=4$). For the distribution-level view, a DeCNN restores feature maps processed by a Re-encoder that shares weights with the Drug Encoder. Finally, the fusion module utilizes multi-head attention ($h=4$) followed by a prediction MLP. Comprehensive hyperparameter details are provided in the Supplementary Material.
\begin{figure}[ht]
    \centering
    \includegraphics[width=0.49\textwidth, trim=52pt 18pt 38pt 25pt, clip]{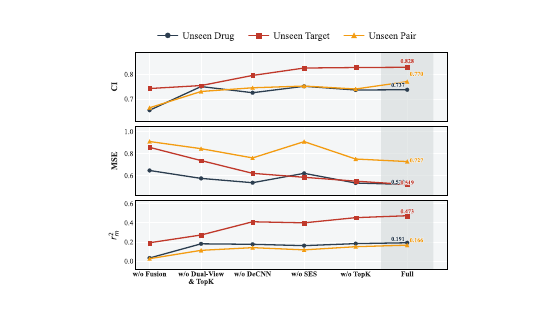}  % 左下右上 0.50 too wide
    \caption{Ablation study on the Davis dataset.}
    %  \textbf{(b)}: Parameter sensitivity analysis of $K$. \textbf{(c)}: Parameter sensitivity analysis of $\lambda$.
    \label{fig:ablation}
\end{figure}
\begin{figure}[ht]
    \centering
    \includegraphics[width=0.49\textwidth, trim=22pt 10pt 20pt 15pt, clip]{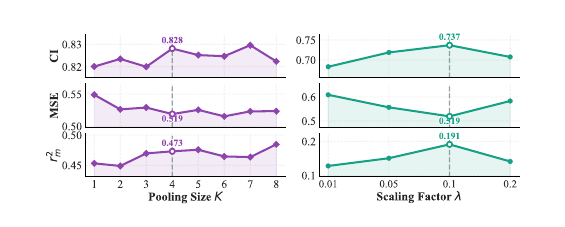}  % 左下右上 0.50 too wide
    \caption{Parameter sensitivity analysis on the Davis dataset.}
    %  \textbf{(b)}: Parameter sensitivity analysis of $K$. \textbf{(c)}: Parameter sensitivity analysis of $\lambda$.
    \label{fig:sensitivity}
\end{figure}

\noindent \textbf{Evaluation Metrics.}
Consistent with \cite{he2025dual}, performance is evaluated using four standard metrics: Mean Absolute Error (\textbf{MAE}) and Mean Squared Error (\textbf{MSE}) quantify the deviation between predicted and actual affinities. The Concordance Index (\textbf{CI}) measures the ranking consistency of the model. Additionally, \textbf{$\mathbf{r^2_m}$}~\cite{pratim2009two} is employed to assess the external predictive capability and reliability of the model on the test sets.

\begin{figure*}[ht]
    \centering
    \includegraphics[width=0.9\textwidth, trim=5pt 10pt 5pt 8pt, clip]{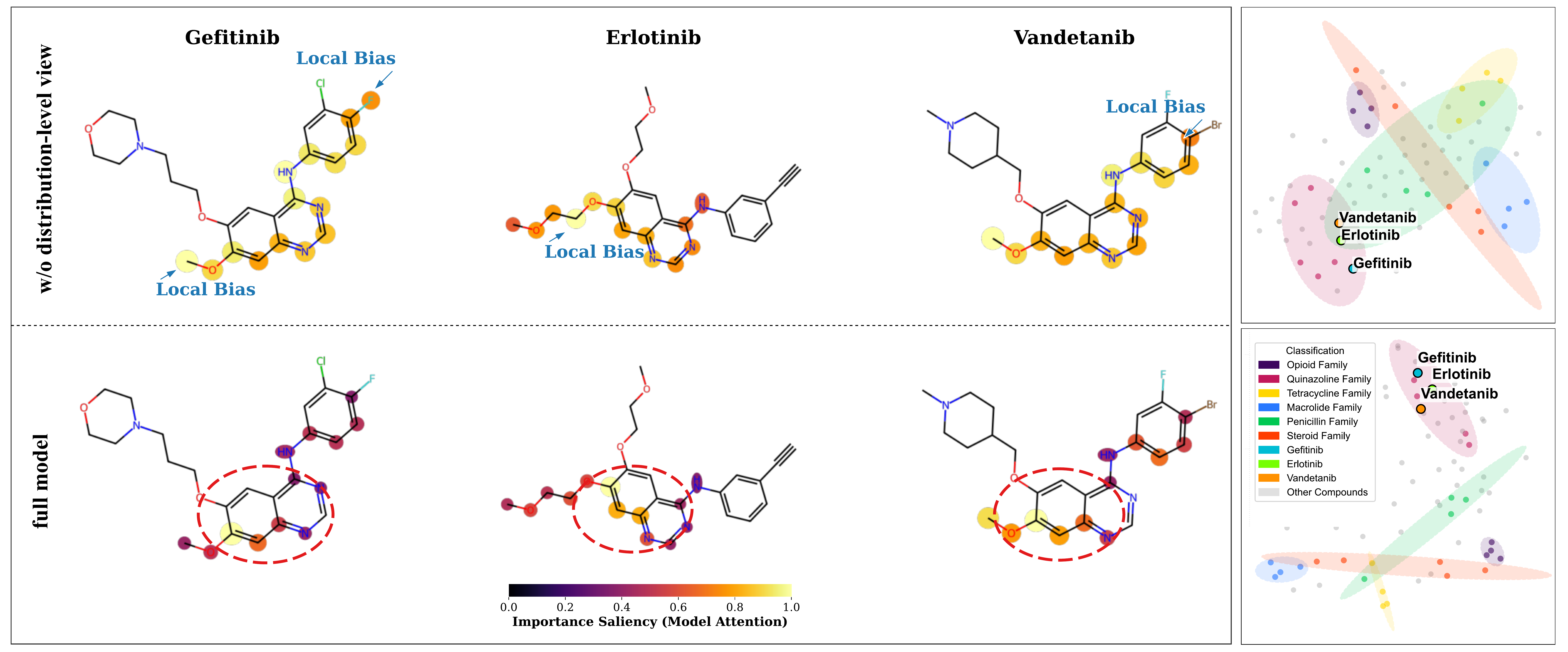}  % 左下右上 0.50 too wide
    \caption{Saliency visualization of molecules with and without distribution-level view integration, and t-SNE projection of learned drug representations. The distinct clustering of drugs sharing similar scaffolds demonstrates that the model effectively captures discriminative structural semantics, mapping chemically similar compounds into proximal latent regions.}\label{fig:fenzikeshihua}
    
\end{figure*}

\noindent \textbf{Baselines}
We benchmark LaPro-DTA against six representative state-of-the-art methods:
DeepDTA~\cite{ozturk2018deepdta} employs dual CNNs for both drug and target encoding;
AttentionDTA~\cite{zhao2019attentiondta} integrates attention mechanisms with CNNs and BiRNNs;
Co-VAE~\cite{li2021co} applies a co-variational autoencoder for joint representation learning;
GraphDTA~\cite{nguyen2021graphdta} processes drug molecular graphs using Graph Isomorphism Networks;
% ColdDTA~\cite{fang2023colddta} focuses on addressing the cold-start problem by learning generalized representations for novel drugs and targets;
TransVAE-DTA~\cite{zhou2024transvae} combines transformers with VAEs in a hybrid framework.
% MultiKD-DTA~\cite{hu2025multikd} uses the pre-trained ESM-2 model~\cite{lin2023evolutionary}, knowledge distillation, and a multiscale feature extraction strategy.
Notably, ColdDTA~\cite{fang2023colddta} performs data augmentation by randomly removing atomic groups from drug molecules to enhance generalization in cold-start scenarios. 
PairVAE~\cite{13} utilizes multi-task learning combined with unsupervised strategies to improve model generalization.
To ensure rigorous fairness, all baselines are \textbf{re-trained} on the experimental datasets, with hyperparameters referenced from the original literature and optimized on the specific experimental data.

\subsection{Performance Comparison and Analysis}
\begin{table}[!ht]
    \centering
    
    \scriptsize % 保持小字体
    \renewcommand{\arraystretch}{0.85} % 保持紧凑行高
    \setlength{\tabcolsep}{10pt} % 适度列宽
    
    \begin{tabular}{lccc}
        \toprule
        \textbf{PDBbind ID} & \textbf{Year} & \textbf{Affinity ($pK_d$)} & \textbf{Pred. Affinity} \\
        \midrule
        5txy & 2017 & 5.460 & 5.476804 \\
        6qlp & 2019 & 5.280 & 5.301821 \\
        5u0e & 2017 & 5.110 & 5.134590 \\
        6i12 & 2019 & 5.570 & 5.541025 \\
        6ma2 & 2018 & 5.389 & 5.419472 \\
        6ajv & 2019 & 5.039 & 5.130577 \\
        \vdots & \vdots & \vdots & \vdots \\ % 省略号占位
        \midrule
        % 底部对比部分
        \multicolumn{4}{c}{\cellcolor{gray!10}\textit{\textbf{Overall Performance (Mean MSE $\downarrow$)}}} \\
        \midrule
        \multicolumn{3}{l}{Pair-VAE} & 1.179 \\
        \multicolumn{3}{l}{HiSIF-DTA (Top-Down)} & 0.978 \\
        \multicolumn{3}{l}{\gc \textbf{LaPro-DTA (Ours)}} & \gc \textbf{0.855} \\
        \bottomrule
    \end{tabular}
    \caption{Predictive performance comparison on recent PDBbind entries (all methods evaluated under \underline{identical conditions}).}
    \label{tab:affinity_prediction}
    
\end{table}
Table~\ref{tab:coldstart} reveals a universal performance decline across all the methods compared to random split benchmarks (details in Supplementary Materials), confirming the inherent difficulty of generalizing to unseen entities.
Despite this challenge, LaPro-DTA consistently establishes new state-of-the-art results across most scenarios. In the \textit{unseen drug} setting, the model demonstrates exceptional generalization, reducing MSE by 8.3\% on Davis and 6.1\% on KIBA relative to the strongest baselines. Similarly, in the \textit{unseen target} scenario, it outperforms best methods by over 6.3\% in terms of MSE on Davis. 
In the most rigorous \textit{unseen pair} scenario, LaPro-DTA maintains competitive dominance. On the Davis dataset, it reduces MSE by 11.3\% and improves $CI$ by 3.9\% compared to PairVAE. On the larger KIBA dataset, while PairVAE yields a marginally lower MSE, our method achieves a superior MAE. 
Notably, unlike PairVAE which relies on a computationally intensive generative framework, LaPro-DTA employs an efficient discriminative architecture.
While PairVAE achieves marginally lower MSE on KIBA, our method strikes a superior balance between accuracy and complexity, demonstrating that mitigating overfitting alone enables robust extrapolation without the heavy computational cost of generative augmentation.
% It is worth noting that unlike PairVAE, which relies on a computationally intensive generative framework, LaPro-DTA utilizes an efficient discriminative architecture. This represents a trade-off between predictive accuracy and model complexity, providing empirical evidence that explicitly mitigating overfitting enables robust extrapolation without the need for heavy generative augmentation.
These results demonstrate the strong generalization capability of the proposed method.

\begin{figure*}[ht]
    \centering
    \includegraphics[width=0.85\textwidth, trim=5pt 10pt 5pt 8pt, clip]{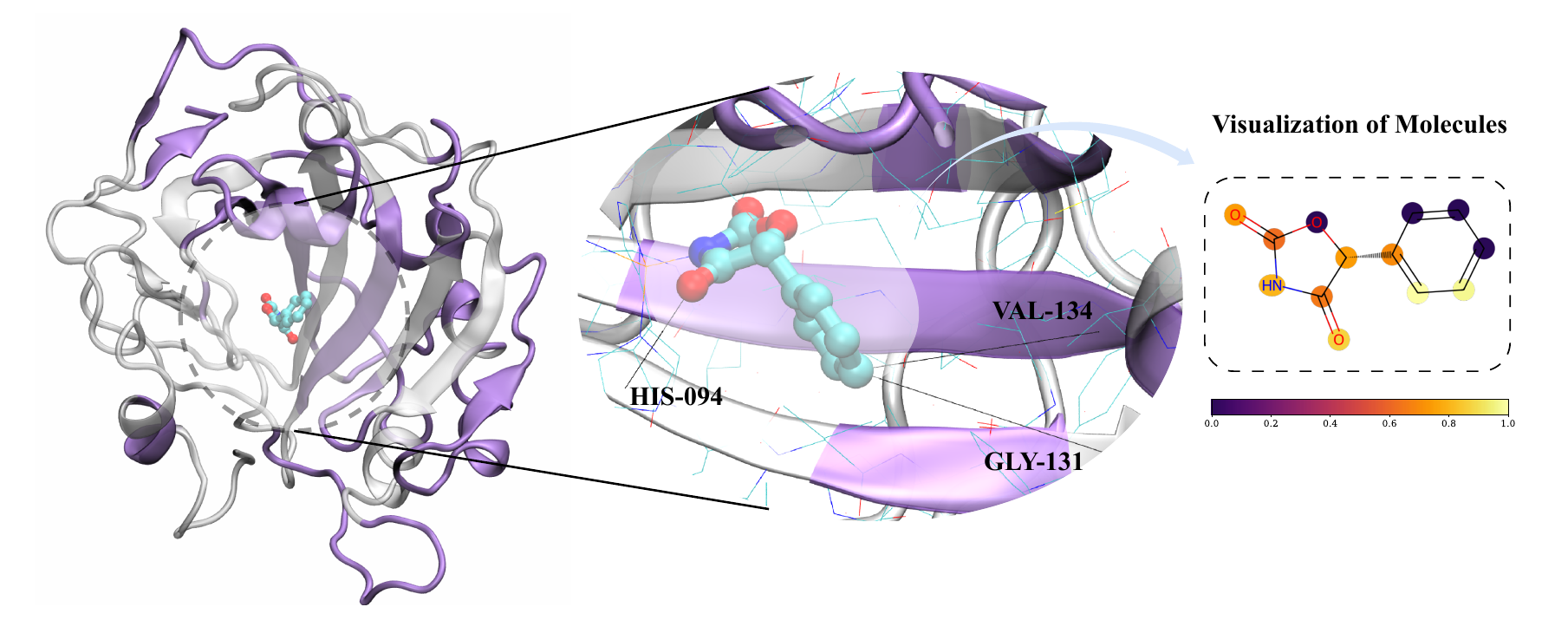}  % 左下右上 0.50 too wide
    \caption{A saliency visualization of the drug--target binding pocket, where colored regions indicate key residues identified by the model.}
    \label{fig:danbai}
\end{figure*}

\begin{table}[!ht]
    \centering
    
    \scriptsize 
    \renewcommand{\arraystretch}{0.8}
    \setlength{\tabcolsep}{4pt} 
    
    \begin{tabular}{c c c l c c c c} 
        \toprule
        \textbf{ID} & 
        \textbf{Year} & 
        \textbf{Structure} & 
        \textbf{SMILES} & 
        \textbf{True} & 
        \textbf{Pred.} & 
        \textbf{MSE} &
        \textbf{Avg.} \\ 
        \midrule

        % ====== Drug 1 (5txy) ======
        \multirow{4}{*}{\textbf{5txy}} & 
        \multirow{4}{*}{2017} & 
        \multirow{4}{*}[-1pt]{\includegraphics[height=0.75cm, keepaspectratio]{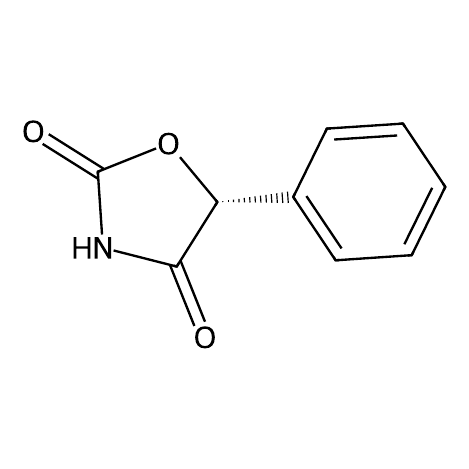}} & 
        \tiny \ttfamily O=C1NC(=...2)O1 & 5.46 & 5.48 & 0.001 &
        \multirow{4}{*}{\textbf{0.036}} \\ 
        
        & & & \tiny \ttfamily O1C(=O)...ccc1 & 5.46 & 5.28 & 0.034 & \\ 
        & & & \tiny \ttfamily O=C1O[C@...ccc1 & 5.46 & 5.20 & 0.067 & \\ 
        & & & \tiny \ttfamily O=C1O[C@...N1 & 5.46 & 5.25 & 0.043 & \\ 
        \midrule

        % ====== Drug 2 (6qlp) ======
        \multirow{4}{*}{\textbf{6qlp}} & 
        \multirow{4}{*}{2019} & 
        \multirow{4}{*}[-1pt]{\includegraphics[height=0.75cm, keepaspectratio]{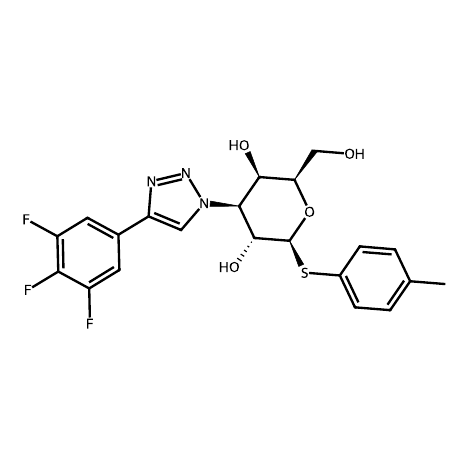}} & 
        \tiny \ttfamily Cc1ccc(...@H]2O & 5.28 & 5.30 & 0.001 &
        \multirow{4}{*}{\textbf{0.007}} \\ 
        
        & & & \tiny \ttfamily Cc1ccc(...)CO & 5.28 & 5.26 & 0.001 & \\ 
        & & & \tiny \ttfamily Cc1ccc(...F)CO & 5.28 & 5.19 & 0.008 & \\ 
        & & & \tiny \ttfamily Cc1ccc(...n1 & 5.28 & 5.14 & 0.019 & \\ 
        \midrule

        % ====== Drug 3 (5u0e) ======
        \multirow{4}{*}{\textbf{5u0e}} & 
        \multirow{4}{*}{2017} & 
        \multirow{4}{*}[-1pt]{\includegraphics[height=0.75cm, keepaspectratio]{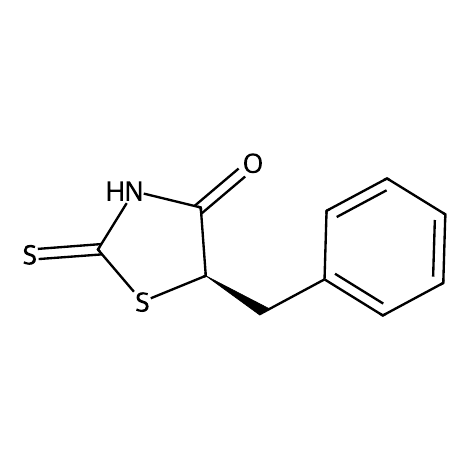}} & 
        \tiny \ttfamily O=C1NC(...ccc1 & 5.11 & 5.13 & 0.001 &
        \multirow{4}{*}{\textbf{0.036}} \\ 
        
        & & & \tiny \ttfamily C([C@@H...ccc1 & 5.11 & 5.42 & 0.096 & \\ 
        & & & \tiny \ttfamily c1cc(C[...c1 & 5.11 & 5.20 & 0.008 & \\ 
        & & & \tiny \ttfamily O=C1[C@...=S & 5.11 & 5.31 & 0.039 & \\ 
        \midrule

        % ====== Drug 4 (6i12) ======
        \multirow{4}{*}{\textbf{6i12}} & 
        \multirow{4}{*}{2019} & 
        \multirow{4}{*}[-1pt]{\includegraphics[height=0.75cm, keepaspectratio]{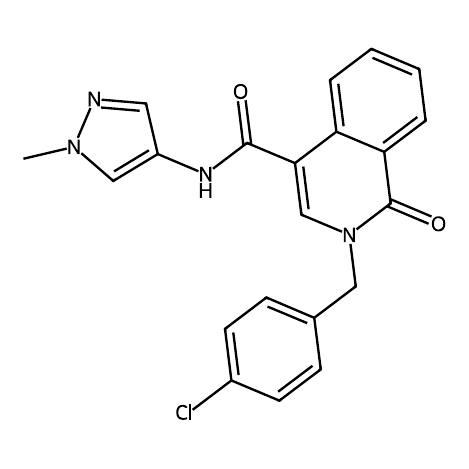}} & 
        \tiny \ttfamily Cn1cc(N...3)cn1 & 5.57 & 5.54 & 0.001 &
        \multirow{4}{*}{\textbf{0.004}} \\ 
        
        & & & \tiny \ttfamily Cn1cc(N...O)cn1 & 5.57 & 5.60 & 0.001 & \\ 
        & & & \tiny \ttfamily Cn1cc(c...21)=O & 5.57 & 5.65 & 0.007 & \\ 
        & & & \tiny \ttfamily Cn1ncc(...c3)c1 & 5.57 & 5.48 & 0.008 & \\ 
        \midrule

        % ====== Drug 5 (6ma2) ======
        \multirow{4}{*}{\textbf{6ma2}} & 
        \multirow{4}{*}{2018} & 
        \multirow{4}{*}[-1pt]{\includegraphics[height=0.75cm, keepaspectratio]{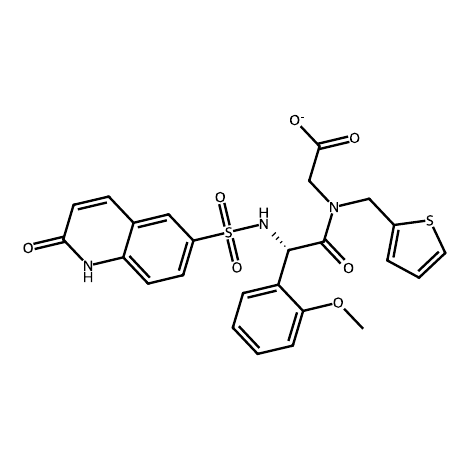}} & 
        \tiny \ttfamily COc1ccc...1cccs1 & 5.39 & 5.42 & 0.001 &
        \multirow{4}{*}{\textbf{0.028}} \\ 
        
        & & & \tiny \ttfamily COc1c([...cc1 & 5.39 & 5.66 & 0.075 & \\ 
        & & & \tiny \ttfamily c1([C@H...c1 & 5.39 & 5.57 & 0.031 & \\ 
        & & & \tiny \ttfamily COc1ccc...H]2)=O & 5.39 & 5.46 & 0.005 & \\ 
        
        \bottomrule
    \end{tabular}
    \caption{Robustness analysis on recent PDBbind entries (SMILES variants). SMILES strings are truncated, denoted by ``...''.}
    \label{tab:smiles_robustness}
    
\end{table}

\subsection{Ablation Study and Parameter Sensitivity}

\noindent \textbf{Effectiveness of Model Components.} 
To validate the specific contribution of each module, we performed a comprehensive ablation study on the Davis dataset, as illustrated in Figure~\ref{fig:ablation}.
First, replacing the cross-view multi-head attention mechanism with simple feature concatenation (w/o Fusion, standard in \cite{ozturk2018deepdta,li2021co}) results in the most severe degradation. This confirms that linear alignment fails to fully integrate the complex interactions between drug and target views.
% First, replacing the \textit{Fusion Block} with simple feature concatenation (standard in \cite{ozturk2018deepdta,li2021co}) results in the most severe degradation. 
% This confirms that our mechanism effectively captures intrinsic binding features, whereas linear alignment fails to fully integrate cross-view representations.
Second, removing the distribution-level view while retaining only the instance-level view (w/o Dual-View) sharply increases error in the unseen pair setting. This supports the hypothesis that the distribution-level view captures transferable chemical scaffolds essential for generalizing to novel entities.
% Second, ablating the \textit{Dual-View Mechanism} (leaving only the instance-level view) sharply increases error in the \textit{Unseen Pair} setting, supporting the distribution-level view's role in capturing transferable features needed for generalizing to novel entities.
Meanwhile, replacing the structure-aware DeCNN module with an equivalent-parameter MLP (w/o DeCNN) degrades performance. This suggests that the inductive bias of convolution is better suited for preserving intrinsic molecular topology than flat projections.
% Meanwhile, replacing DeCNN with an equivalent-parameter MLP degrades performance. This suggests that DeCNN is better suited for retaining the intrinsic chemical structure of drug features than simple linear layers.
Finally, substituting pattern-aware top-$k$ pooling with standard max pooling (w/o TopK, as in \cite{zhou2024transvae,zhang2024gdilateddta}) leads to a notable decline. This validates that top-$k$ strategy effectively preserves critical bio-active residues and mitigates information loss caused by non-essential background sequences.
These components function synergistically to secure optimal performance.
% To validate the contribution of each module, we performed an ablation study on the Davis dataset (Figure~\ref{fig:ablation}(a)). 
% Removing the \textbf{Fusion Block} results in the most severe degradation, confirming that simple feature concatenation is insufficient for aligning heterogeneous dual-view features; the attention-based interaction is essential for synthesizing comprehensive representations. 
% Notably, ablating the \textbf{Dual-View Mechanism} leads to a sharp error increase in the \textit{Unseen Pair} setting. This supports our hypothesis that the Global-Context view is indispensable for capturing transferable scaffolds when reasoning about novel entities.
% Furthermore, the performance drop observed in the \textit{-Deconv} variant (replacing DeCNN with an MLP) verifies that the gain stems from the architectural inductive bias of convolutions—which enforce spatial continuity—rather than mere parameter expansion.
% Finally, replacing \textbf{Top-K Pooling} with Max Pooling impairs performance, demonstrating that retaining a spectrum of bioactive patterns is superior to single-value abstraction for filtering background noise. 
% Collectively, these results affirm that the components function synergistically to secure robust prediction against cold-start challenges.

\noindent \textbf{Parameter Sensitivity.} 
We evaluate the impact of $K$ and $\lambda$ in Figure~\ref{fig:sensitivity}. 
Performance peaks at $K=4$; larger values introduce background noise and cause \textbf{information loss}, validating the sparsity assumption of binding pockets. 
Similarly, $\lambda=0.1$ yields the best results; lower values neglect global semantics while higher ones overwhelm instance's local details, confirming $0.1$ as the optimal balance for dual-view fusion.

\subsection{Effectiveness of Distribution-Level View}

To validate the capability of the distribution-level view in identifying drug scaffolds, we compared attention maps for structurally similar drugs, using Gefitinib as a case study (Figure~\ref{fig:fenzikeshihua}). The baseline model relies solely on the instance-level view and exhibits structural myopia, causing it to overfit to spurious peripheral atoms. In contrast, the distribution-level view serves as a topological regularizer, redirecting focus toward the conserved main skeleton and key pharmacophores. This effect is further confirmed by t-SNE visualization~\cite{vandermaaten2008visualizing}, where related compounds show sharper clustering. These results demonstrate that the proposed synergy effectively corrects local biases by enforcing holistic structural constraints.

\subsection{Robustness Analysis}
\noindent \textbf{Invariance to SMILES Permutations.}
To evaluate the generalization capability of LaPro-DTA, we selected recent entries from the PDBbind dataset (as shown in Table~\ref{tab:smiles_robustness}). Despite variations in SMILES linearization orders (generated by RDKit), LaPro-DTA exhibits remarkable stability. This observation indicates that the SES design effectively mitigates the model’s reliance on shortcuts inherent in SMILES syntax, enabling a clear decoupling between molecular semantics and syntactic permutations, and thereby preserving intrinsic molecular semantics.
\vspace{-6pt}

\subsection{Out-of-Sample Test}

% \noindent \textbf{Dataset and Setup.}
To rigorously evaluate generalization under a strict and impartial out-of-distribution setting, we adopt the external validation protocol from HiSIF-DTA~\cite{bi2023hisif}.
To fairly assess zero-shot transferability, all compared methods utilize models trained solely on the Davis dataset. These models are evaluated on 95 fresh PDBbind pairs that have zero overlap with the Davis dataset, ensuring no data leakage (see Supplementary Materials for details).
% \noindent \textbf{Results and Analysis.}
As detailed in Table~\ref{tab:affinity_prediction}, LaPro-DTA achieves an MSE of \textbf{0.855}, significantly outperforming the HiSIF-DTA (12\%) under identical conditions. These results confirm that our model effectively captures transferable physicochemical semantics rather than dataset-specific biases, enabling robust generalization to realistic molecular distributions beyond standard benchmarks.

\subsection{Interpretability of Protein Feature Extraction}
\noindent \textbf{Binding Interface Localization.}
Figure~\ref{fig:danbai} illustrates the interaction for the representative complex PDB 5txy. 
We conducted this visualization analysis using the Grad-CAM technique~\cite{selvaraju2017grad} and VMD software~\cite{humphrey1996vmd}. 
The pattern-aware top-$k$ Pooling effectively filters non-essential protein regions, concentrating high attention scores strictly around the actual binding pocket. Notably, critical contacting residues—including HIS-094, GLY-131, and VAL-134—are identified. This spatial alignment confirms that LaPro-DTA prioritizes physical interaction interfaces over irrelevant background sequences.

\section{Conclusion}
In this paper, we proposed LaPro-DTA to tackle the critical generalization bottleneck in cold-start DTA prediction.
It effectively mitigates overfitting through a latent dual-view drug representation mechanism and alleviates information loss via a pattern-aware salient target feature extraction strategy.
Extensive experiments demonstrate that LaPro-DTA not only establishes new state-of-art performance in challenging unseen settings but also offers interpretability into the underlying binding mechanisms.
For future work, we plan to extend our framework beyond 1D sequences by integrating explicit geometric graphs or 3D conformers, aiming to unlock richer spatial dependencies for even more robust drug discovery.

\newpage

\appendix

% This supplementary material presents a comprehensive exposition of the experimental protocols, additional empirical results, and performance evaluations of the proposed LaPro-DTA under the standard random-split setting. 
% Specifically, we provide an extended review of related work, algorithmic details of the cold-start splitting strategy, and extensive benchmarks on the Davis~\cite{davis2011comprehensive}, KIBA~\cite{tang2014making}, and BindingDB~\cite{liu2007bindingdb} datasets. 
% Through rigorous comparative analysis, we further corroborate the efficacy of our method. 
% Finally, to facilitate reproducibility, we release the full source code for LaPro-DTA, alongside the retraining scripts for the GraphDTA~\cite{nguyen2021graphdta} baseline adapted for cold-start scenarios.

\section{Related Works}

% 现有的药物-靶点亲和力（DTA）预测计算框架通常可以解构为三个主要组成部分：药物表征建模模块、靶点表征建模模块以及特征融合与预测模块。因此，在本节中，我们将遵循这一分类体系，对相关文献进行系统的回顾与梳理。
\noindent \textbf{Overview.}
Contemporary computational frameworks for DTA prediction generally comprise three core components: drug representation learning, target representation learning, and feature fusion. 
To articulate the motivation underlying LaPro-DTA, we systematically review the literature through the lens of this taxonomy.
\subsection{Drug Representation Modeling}
Pioneering studies, such as DeepDTA~\cite{ozturk2018deepdta} and WideDTA~\cite{ozturk2019widedta}, conceptualize SMILES strings as one-dimensional sequences, employing Convolutional Neural Networks (CNNs) or Recurrent Neural Networks (RNNs) to extract local patterns. 
Despite their efficacy, these sequence-based methods are prone to overfitting local lexical features, often failing to capture the intrinsic spatial geometry of molecules. 
To mitigate this limitation, GraphDTA~\cite{nguyen2021graphdta} represents drug molecules as graphs, leveraging Graph Neural Networks (GNNs) to explicitly encode topological dependencies. 
Building on this paradigm, GDilatedDTA~\cite{zhang2024gdilateddta} synergizes 2D molecular graph features with 1D sequence features, utilizing a hybrid GNN-CNN architecture to capture complementary multimodal information. 
Alternatively, Co-VAE~\cite{li2021co} introduces a multi-task learning framework based on Variational Autoencoders (VAEs); by incorporating molecular reconstruction as an auxiliary objective, it facilitates the learning of regularized and semantic-rich latent drug representations.

\subsection{Target Representation Modeling}
% 靶点表征建模 由于氨基酸序列显著的长度和丰富的信息量，蛋白质表征建模相比药物分子面临着更大的挑战。为了提取局部序reconstructs the molec列特征，CNN 和 RNN 被广泛应用，代表性工作包括 DeepDTA、WideDTA 和 AttentionDTA。此外，Co-VAE 将 VAE 框架扩展至蛋白质建模，旨在学习更紧凑的潜在表征。更近期的 TransVAE-DTA 则利用 Transformer 编码器来捕获长距离依赖关系以提升全局建模能力。然而，值得注意的是，在特征提取阶段之后，现有的绝大多数方法主要依赖简单的全局最大池化或平均池化，将整个变长序列聚合为单一的固定长度向量。
Target representation modeling poses a more formidable challenge than drug encoding, primarily attributable to the extensive length and high information density of amino acid sequences. 
To capture local sequential motifs, Convolutional and Recurrent Neural Networks are widely adopted, as exemplified by DeepDTA~\cite{ozturk2018deepdta}, WideDTA~\cite{ozturk2019widedta}, and AttentionDTA~\cite{zhao2019attentiondta}. 
Advancing beyond standard architectures, Co-VAE~\cite{li2021co} adapts the VAE framework to distill compact latent representations, whereas TransVAE-DTA~\cite{zhou2024transvae} leverages Transformer encoders to model long-range dependencies and global context. 
Crucially, despite these advances in feature extraction, the prevailing paradigm relies disproportionately on rudimentary global pooling (\textit{e.g.}, max or average pooling) to compress variable-length sequences into fixed-size vectors.

\subsection{Feature Fusion Module}
The feature fusion module serves as the nexus for synthesizing drug and target representations to predict binding affinity. 
Seminal models like DeepDTA~\cite{ozturk2018deepdta} relied on the rudimentary concatenation of global embeddings, a strategy that inevitably neglects fine-grained intermolecular interactions. 
To capture substructure-level dependencies, MolTrans~\cite{huang2021moltrans} constructs pairwise interaction maps between drug substructures and protein fragments, explicitly modeling high-order interaction patterns. 
Subsequently, attention-driven paradigms further advanced interaction modeling: AttentionDTA~\cite{zhao2019attentiondta} employs cross-attention mechanisms to isolate relevant interaction regions, whereas TransVAE-DTA~\cite{zhou2024transvae} leverages Attention-based Pooling (AAP) for more discriminative feature aggregation. 
However, owing to severe compression of target features in upstream encoding, these attention mechanisms are prone to degeneration, significantly undermining their efficacy in capturing fine-grained bioactivity.

% 尽管现有方法取得了一定的成功，但它们仍存在显著的局限性。具体而言，药物和蛋白质编码器通常依赖于单一视角的嵌入，往往忽略了显式的药物全局结构特征以及关键的蛋白质局部片段信息。此外，现有的融合模块——无论是基于简单的拼接、交互映射还是注意力机制——普遍缺乏增强表征鲁棒性或对齐互补特征的有效机制。这些缺陷共同导致了模型在面对未见药物或蛋白质时泛化能力受限。
Despite the success achieved by existing methods, significant limitations persist. 
Specifically, drug and protein encoders typically rely on single-view embeddings, often overlooking explicit distribution perspectives for drugs and key local fragment information for proteins. 
Moreover, fusion modules—whether based on concatenation, interaction maps, or attention—generally lack mechanisms to enhance robustness or align complementary features. 
Consequently, these deficiencies severely limit the generalization capabilities of current models when encountering unseen drugs or targets.

\section{Supplementary Experiments}
\subsection{More Implementation Details}
The detailed hyperparameter configurations are summarized in Table \ref{tab:hyperparameters}. We implemented LaPro-DTA using the PyTorch 2.0 framework on an NVIDIA GeForce RTX 3090 GPU. To prevent information leakage and simulate realistic cold-start scenarios, we applied a structured data splitting protocol (Algorithm \ref{alg:data_split_oog}). Let $\mathcal{D}$ and $\mathcal{P}$ denote the sets of drugs and proteins, respectively. We randomly partitioned these sets into disjoint subsets: $\mathcal{D} = \mathcal{D}_{seen} \cup \mathcal{D}_{unseen}$ and $\mathcal{P} = \mathcal{P}_{seen} \cup \mathcal{P}_{unseen}$.

\begin{algorithm}[ht]
\SetAlgoLined
\caption{Cold-Start Data Splitting Algorithm}
\label{alg:data_split_oog}

\KwIn{Interaction Matrix $\mathbf{A}$, Unseen Ratio $\rho$ }
\KwOut{Train Set $\mathcal{D}_{train}$, Val Set $\mathcal{D}_{val}$, Test Sets $\{\mathcal{D}_{S2}, \mathcal{D}_{S3}, \mathcal{D}_{S4}\}$}

\BlankLine
\tcc{Step 1: Entity-Level Partitioning}
Set random seed $s$\;
Identify active drugs $\mathcal{I}_d$ and proteins $\mathcal{I}_p$ from $\mathbf{A}$\;
$K_d \leftarrow \lfloor \rho \times |\mathcal{I}_d| \rfloor$, $K_p \leftarrow \lfloor \rho \times |\mathcal{I}_p| \rfloor$\;
Randomly select \textbf{unseen} sets $\mathcal{U}_d \subset \mathcal{I}_d$, $\mathcal{U}_p \subset \mathcal{I}_p$ of size $K_d, K_p$\;
Define \textbf{seen} sets $\mathcal{S}_d \leftarrow \mathcal{I}_d \setminus \mathcal{U}_d$, $\mathcal{S}_p \leftarrow \mathcal{I}_p \setminus \mathcal{U}_p$\;

\BlankLine
\tcc{Step 2: Interaction Allocation}
Initialize $\mathcal{D}_{train}, \mathcal{D}_{val}, \mathcal{D}_{S2}, \mathcal{D}_{S3}, \mathcal{D}_{S4} \leftarrow \emptyset$\;

\For{each interaction $(d, p)$ such that $\mathbf{A}_{d,p} > 0$}{
    \uIf{$d \in \mathcal{S}_d$ \textbf{and} $p \in \mathcal{S}_p$}{
        \tcp{Seen Drug, Seen Protein}
        Add $(d, p)$ to $\mathcal{D}_{train}$\;
    }
    \Else{
        Generate random $r \sim Uniform(0, 1)$\;
        \uIf{$d \in \mathcal{U}_d$ \textbf{and} $p \in \mathcal{S}_p$}{
            \tcp{New Drug (S2)}
            \lIf{$r < 0.5$}{Add to $\mathcal{D}_{val}$}
            \lElse{Add to $\mathcal{D}_{S2}$}
        }
        \uElseIf{$d \in \mathcal{S}_d$ \textbf{and} $p \in \mathcal{U}_p$}{
            \tcp{New Protein (S3)}
            \lIf{$r < 0.5$}{Add to $\mathcal{D}_{val}$}
            \lElse{Add to $\mathcal{D}_{S3}$}
        }
        \ElseIf{$d \in \mathcal{U}_d$ \textbf{and} $p \in \mathcal{U}_p$}{
            \tcp{Both New (S4)}
            \lIf{$r < 0.5$}{Add to $\mathcal{D}_{val}$}
            \lElse{Add to $\mathcal{D}_{S4}$}
        }
    }
}
\Return $\mathcal{D}_{train}, \mathcal{D}_{val}, \mathcal{D}_{S2}, \mathcal{D}_{S3}, \mathcal{D}_{S4}$\;
\end{algorithm}

\begin{table}[ht]
    \centering
    \caption{Hyperparameter settings of LaPro-DTA.}
    \label{tab:hyperparameters}
    
    % 使用 small 字体，不再强制拉伸到 \linewidth
    \small 
    % 稍微增加行高，提升阅读舒适度
    \renewcommand{\arraystretch}{1.15}
    % 稍微增加列间距，避免太挤，但不会撑满
    \setlength{\tabcolsep}{10pt}
    
    \begin{tabular}{l l}
        \toprule
        \textbf{Hyperparameter} & \textbf{Value} \\
        \midrule
        
        % --- Drug ---
        \multicolumn{2}{l}{\textit{\textbf{Drug Representation}}} \\
        Embedding Dimension & 128 \\
        CNN Filter Sizes & [4, 4, 4] \\
        Number of Filters & 128 \\
        DeconvNet Layers & 3 \\
        Latent Dimension ($z$) & 96 \\
        \midrule
        
        % --- Target ---
        \multicolumn{2}{l}{\textit{\textbf{Target Representation}}} \\
        Embedding Dimension & 128 \\
        CNN Filter Sizes & [4, 8, 12] \\
        Top-$K$ Pooling Size ($K$) & 4 \\
        Number of CNN Layers & 3 \\
        \midrule
        
        % --- Interaction ---
        \multicolumn{2}{l}{\textit{\textbf{Interaction Module}}} \\
        Attention Heads & 4 \\
        Attention Hidden Dim ($d_k$) & 96 \\
        MLP Hidden Units & [1024, 512] \\
        \midrule
        
        % --- Optimization ---
        \multicolumn{2}{l}{\textit{\textbf{Optimization}}} \\
        Optimizer & Adam \\
        Learning Rate & $5 \times 10^{-4}$ \\
        Batch Size (KIBA / Davis) & 256 / 32 \\
        Max Epochs (Cold / Rand) & 100 / 500 \\
        Dropout Rate & 0.1 \\
        Weight Decay & $1 \times 10^{-4}$ \\
        \bottomrule
    \end{tabular}
\end{table}

\noindent  \textbf{Datasets.} 
We evaluate LaPro-DTA on two standard benchmarks, \textbf{Davis}~\cite{davis2011comprehensive} and \textbf{KIBA}~\cite{tang2014making}, while incorporating \textbf{BindingDB}~\cite{liu2007bindingdb} to assess large-scale generalization. Davis contains 30,056 interactions ($pK_d$) between 68 drugs and 442 targets, whereas KIBA comprises 118,254 entries (KIBA scores) across 2,111 drugs and 229 targets. BindingDB serves as a supplementary testbed with broader chemical diversity.

\begin{figure*}[t]
    \centering
    \includegraphics[width=\textwidth, trim=5pt 5pt 5pt 25pt, clip]{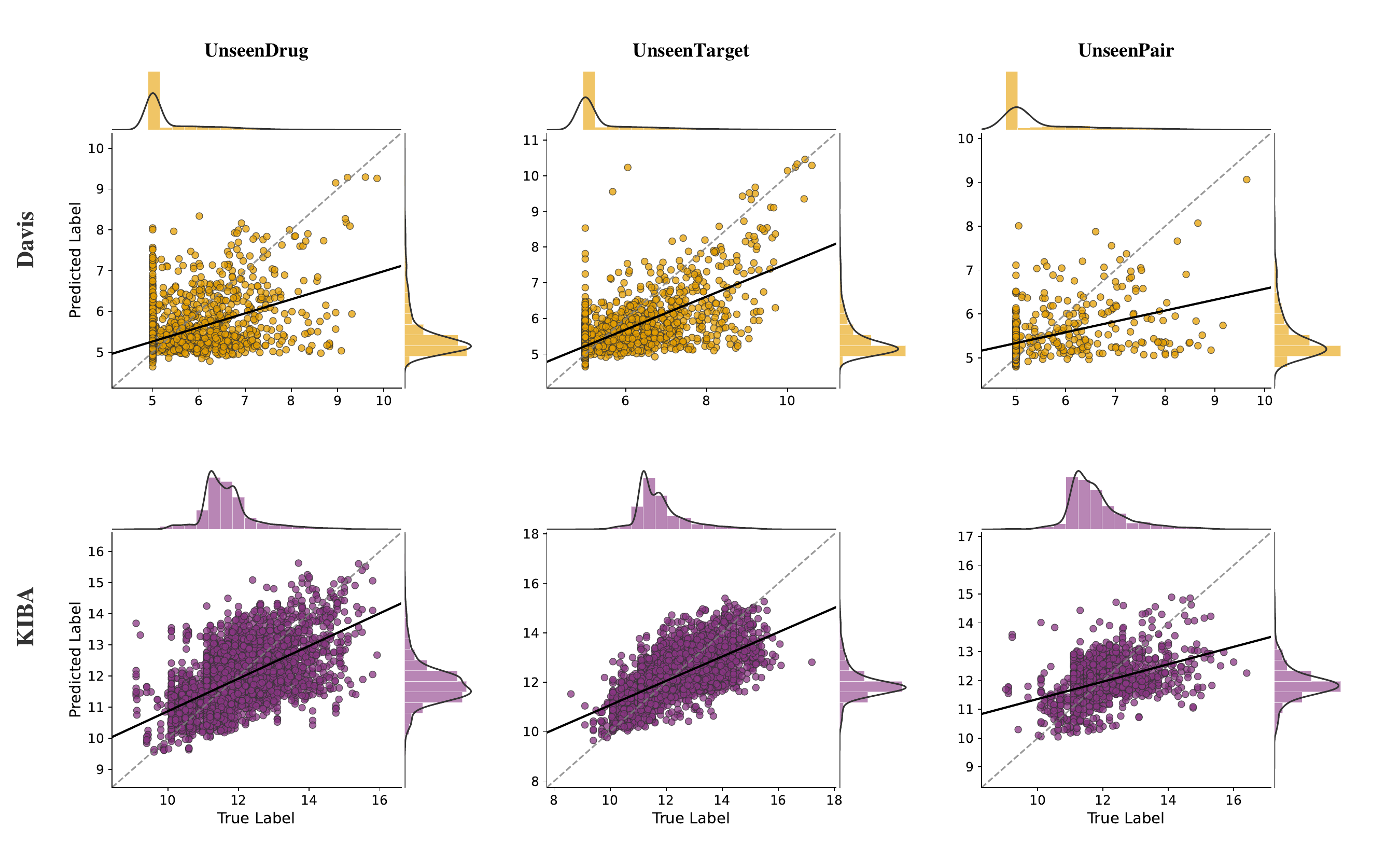}  % 左下右上 0.50 too wide
    \caption{True affinity value against the predicted value on Davis and KIBA datasets.}
    \label{fig:Scatter}
\end{figure*}

\subsection{Experiments on Random-Split Setting}

\noindent \textbf{Supplementary Baselines.} 
To comprehensively validate the effectiveness of LaPro-DTA across different evaluation scenarios, we supplement our analysis by comparing it against a broader range of methods under the standard random split setting. The performance results for the following baselines are directly cited from their respective original publications:
\begin{table*}[!ht]
    \centering
    \caption{Performance comparison of representative DTA models on KIBA, Davis, and BindingDB datasets. (\textbf{Best}, \underline{Second Best}).}
    \label{tab:all_comparison}
    % \small % 如果表格依然太宽，可以取消注释这一行
    
    \resizebox{\textwidth}{!}{
        % 移除了中间的 ll (Architecture列)，现在结构更紧凑
        \begin{tabular}{l | ccc | ccc | ccc}
            \toprule
            \multirow{2}{*}{\textbf{Model}} 
            & \multicolumn{3}{c}{\textbf{KIBA Dataset}} 
            & \multicolumn{3}{c}{\textbf{Davis Dataset}} 
            & \multicolumn{3}{c}{\textbf{BindingDB Dataset}} \\
            
            % 调整横线跨度：现在的列号变了
            \cmidrule(lr){2-4} \cmidrule(lr){5-7} \cmidrule(lr){8-10}
            
            & \textbf{MSE}$\downarrow$ & \textbf{CI}$\uparrow$ & $\mathbf{r_m^2}$$\uparrow$ 
            & \textbf{MSE}$\downarrow$ & \textbf{CI}$\uparrow$ & $\mathbf{r_m^2}$$\uparrow$
            & \textbf{MSE}$\downarrow$ & \textbf{CI}$\uparrow$ & $\mathbf{r_m^2}$$\uparrow$ \\
            \midrule
            
            KronRLS (2015)
            & 0.411 & 0.782 & 0.342 
            & 0.379 & 0.871 & 0.407 
            & 0.910 & 0.710 & -- \\

            SimBoost (2017)
            & 0.222 & 0.836 & 0.629 
            & 0.282 & 0.872 & 0.644 
            & -- & -- & -- \\
            
            DeepDTA (2018)
            & 0.194 & 0.863 & 0.630 
            & 0.261 & 0.878 & 0.630 
            & 0.633 & 0.844 & 0.633 \\

            WideDTA (2019)
            & 0.179 & 0.875 & -- 
            & 0.262 & 0.886 & --
            & -- & -- & -- \\
            
            AttentionDTA (2022)
            & 0.155 & 0.882 & 0.755 
            & \textbf{0.216} & 0.887 & 0.677 
            & 0.745 & 0.542 & -- \\

            GraphDTA (GIN) (2021)
            & \underline{0.147} & \textbf{0.891} & 0.687 
            & 0.229 & \underline{0.893} & 0.663 
            & 0.535 & 0.858 & -- \\

            DeepCDA (2020)
            & 0.176 & 0.889 & 0.682
            & 0.248 & 0.891 & 0.649 
            & 0.848 & 0.722 & 0.531 \\

            Co-VAE (New-drug) (2022)
            & 0.416 & 0.742 & 0.358
            & 0.724 & 0.712 & 0.107 
            & -- & -- & -- \\

            Co-VAE (New-target) (2022)
            & 0.421 & 0.741 & 0.356
            & 0.419 & 0.816 & 0.477
            & -- & -- & -- \\
            
            DoubleSG-DTA (2023)
            & 0.164 & 0.856 & 0.721 
            & 0.250 & 0.886 & \underline{0.688} 
            & 0.533 & 0.862 & \underline{0.726} \\
            
            GDilatedDTA (2024)
            & 0.156 & 0.876 & \textbf{0.775} 
            & 0.237 & 0.885 & 0.686 
            & \underline{0.483} & \underline{0.868} & \textbf{0.730} \\
            
            \midrule
            
            \gc \textbf{LaPro-DTA (Ours)} 
            & \gc \textbf{0.133} & \gc \textbf{0.891} & \gc \underline{0.760} 
            & \gc \underline{0.225} & \gc \textbf{0.895} & \gc \textbf{0.693} 
            & \gc \textbf{0.466} & \gc \textbf{0.874} & \gc 0.706 \\
            \bottomrule
        \end{tabular}
    }
\end{table*}

\begin{table*}[!ht]
    \centering
    % 1. 增加行高，让表格看起来更舒展
    \renewcommand{\arraystretch}{1.2}
    
    \caption{Ablation study of LaPro-DTA on the Davis dataset under three cold-start settings. \textbf{Dual}: Latent Dual-View Mechanism; \textbf{TopK}: top-$k$ Pooling; \textbf{Fusion}: Cross-View Attention. The best results are highlighted in \textbf{bold}, and the second-best results are \underline{underlined}. ``$\uparrow$'' indicates higher is better, ``$\downarrow$'' indicates lower is better.}
    \label{tab:ablation_davis_full_rm2}
    
    % 2. 字体调整为 small，比 footnotesize 更清晰
    \small 
    
    % 3. 使用 tabular* 撑满 \textwidth，利用 @{\extracolsep{\fill}} 自动分配列间距
    \begin{tabular*}{\textwidth}{@{\extracolsep{\fill}} ccc ccc ccc ccc }
        \toprule
        \multicolumn{3}{c}{\textbf{Components}} & \multicolumn{3}{c}{\textbf{Unseen Drug}} & \multicolumn{3}{c}{\textbf{Unseen Target}} & \multicolumn{3}{c}{\textbf{Unseen Pair}} \\
        
        % 调整 cmidrule 跨度，并在中间留出间隙
        \cmidrule(r){1-3} \cmidrule(lr){4-6} \cmidrule(lr){7-9} \cmidrule(l){10-12}
        
        \textbf{Dual} & \textbf{TopK} & \textbf{Fusion} 
        & \textbf{CI} ($\uparrow$) & \textbf{MSE} ($\downarrow$) & $\mathbf{r_m^2}$ ($\uparrow$) 
        & \textbf{CI} ($\uparrow$) & \textbf{MSE} ($\downarrow$) & $\mathbf{r_m^2}$ ($\uparrow$) 
        & \textbf{CI} ($\uparrow$) & \textbf{MSE} ($\downarrow$) & $\mathbf{r_m^2}$ ($\uparrow$) \\
        \midrule
        
        % --- Baseline (都不加) ---
        -- & -- & -- 
        & 0.723 & 0.567 & 0.137 
        & \underline{0.808} & 0.654 & 0.368 
        & 0.728 & 0.836 & 0.104 \\
        
        % --- 变体 1: 去掉 TopK ---
        \checkmark & -- & \checkmark 
        & \underline{0.736} & \underline{0.532} & 0.181 
        & 0.798 & \underline{0.548} & \underline{0.453} 
        & \underline{0.739} & \underline{0.750} & \underline{0.150} \\
        
        % --- 变体 2: 去掉 Dual ---
        -- & \checkmark & \checkmark 
        & 0.726 & 0.542 & \textbf{0.201} 
        & 0.755 & 0.752 & 0.302 
        & \underline{0.739} & 0.804 & 0.124 \\
        
        % --- 变体 3: 去掉 Fusion ---
        \checkmark & \checkmark & -- 
        & 0.654 & 0.646 & 0.032 
        & 0.742 & 0.856 & 0.191 
        & 0.664 & 0.909 & 0.025 \\
        \midrule
        
        % --- Full Model (完整版) ---
        \checkmark & \checkmark & \checkmark 
        & \textbf{0.737} & \textbf{0.519} & \underline{0.191} 
        & \textbf{0.828} & \textbf{0.519} & \textbf{0.473} 
        & \textbf{0.770} & \textbf{0.727} & \textbf{0.166} \\
        \bottomrule
    \end{tabular*}
\end{table*}

\begin{table}[ht]
    \centering
    \caption{Ablation study on the Davis dataset under the \textbf{random split} setting. \textbf{Dual}: latent dual-view (only instance-level view); \textbf{TopK}: top-$k$ pooling; \textbf{Fusion}: cross-view attention.}
    \label{tab:ablation_random_split}
    
    % 使用 resizebox 确保表格完美适应单栏宽度
    \resizebox{\linewidth}{!}{
        \begin{tabular}{ccc|ccc}
            \toprule
            \multicolumn{3}{c|}{\textbf{Components}} & \multicolumn{3}{c}{\textbf{Metrics}} \\
            \cmidrule(r){1-3} \cmidrule(l){4-6} 
            
            \textbf{Dual} & \textbf{TopK} & \textbf{Fusion} 
            & \textbf{CI} ($\uparrow$) & \textbf{MSE} ($\downarrow$) & $\mathbf{r_m^2}$ ($\uparrow$) \\
            \midrule
            
            % Row 1: Baseline
            $\times$ & $\times$ & $\times$ 
            & 0.854 & 0.347 & 0.559 \\
            
            % Row 2: TopK only
            $\times$ & \checkmark & $\times$ 
            & 0.875 & 0.240 & \underline{0.663} \\
            
            % Row 3: Fusion only
            $\times$ & $\times$ & \checkmark 
            & \underline{0.885} & 0.262 & 0.598 \\
            
            % Row 4: TopK + Fusion
            $\times$ & \checkmark & \checkmark 
            & 0.881 & 0.242 & 0.651 \\
            
            % Row 5: Dual + Fusion
            \checkmark & $\times$ & \checkmark 
            & 0.880 & 0.258 & 0.607 \\
            
            % Row 6: Dual + TopK
            \checkmark & \checkmark & $\times$ 
            & 0.882 & \underline{0.239} & 0.641 \\
            \midrule
            
            % Row 7: Full Model
            \checkmark & \checkmark & \checkmark 
            & \textbf{0.895} & \textbf{0.225} & \textbf{0.693} \\
            \bottomrule
        \end{tabular}
    }
\end{table}

\begin{itemize}
    \item \textbf{KronRLS}~\cite{nascimento2016multiple} and \textbf{SimBoost}~\cite{he2017simboost} serve as representative traditional machine learning baselines. KronRLS computes structural and sequence similarity kernels and employs Kronecker regularized least squares for prediction. SimBoost utilizes gradient boosting machines combined with feature engineering to model drug-target pairs.
    
    \item \textbf{DeepDTA}~\cite{ozturk2018deepdta} is a pioneering deep learning framework that treats drug SMILES and protein amino acid sequences as 1D text data, employing dual Convolutional Neural Networks (CNNs) to extract local latent patterns.
    
    \item \textbf{WideDTA}~\cite{ozturk2019widedta} improves upon DeepDTA by adopting a "word-based" approach. Instead of character-level inputs, it utilizes Ligand Maximum Common Substructures (LMCS) and protein motifs as input words for CNNs to capture higher-level semantic features.
    
    \item \textbf{AttentionDTA}~\cite{zhao2019attentiondta} incorporates attention mechanisms into the sequence modeling process. By weighting the interaction intensity between drug and protein subsequences, it aims to focus the model on the most relevant binding regions.
    
    \item \textbf{GraphDTA}~\cite{nguyen2021graphdta} represents a significant shift towards geometric deep learning. It models drug molecules as 2D graphs and investigates various Graph Neural Networks (including GCN, GAT, and GIN) to explicitly capture the topological dependencies of atoms, achieving strong baseline performance.
    
    \item \textbf{DeepCDA}~\cite{abbasi2020deepcda} combines CNNs with Long Short-Term Memory (LSTM) networks. It is designed to capture both local features and long-term sequential dependencies, emphasizing effective encoding for cross-domain affinity prediction.
    
    \item \textbf{Co-VAE}~\cite{li2021co} employs a multi-task learning strategy based on Variational Autoencoders (VAEs). It generates latent representations by jointly optimizing for affinity prediction and the reconstruction of molecular sequences.
    
    \item \textbf{DoubleSG-DTA}~\cite{qian2023doublesg} introduces a multi-view fusion strategy. It extracts features from both the SMILES sequence (via CNN) and the molecular graph (via GNN), integrating these dual representations to enhance drug encoding.
    
    \item \textbf{GDilatedDTA}~\cite{zhang2024gdilateddta} addresses the issue of long-range dependencies and signal loss. It utilizes dilated graph convolutions for drugs and dilated 1D convolutions for proteins to expand the receptive field without increasing model complexity.
\end{itemize}
Table~\ref{tab:all_comparison} reports the performance of LaPro-DTA against representative baselines under the random-split setting, following the protocol of \cite{zhang2024gdilateddta}. Collectively, these experimental results validate the superiority of our proposed method across multiple benchmark datasets.

\subsubsection{Comparative Results and Analysis}
The empirical results presented in Table~\ref{tab:all_comparison} demonstrate that LaPro-DTA establishes new state-of-the-art (SOTA) performance across the majority of evaluation scenarios.

On the KIBA dataset, our model achieves an optimal MSE of 0.133 and a CI of 0.891. Notably, this performance significantly surpasses not only the 1D-based DeepDTA but also the 2D graph-based GraphDTA (which yields an MSE of 0.147. This substantial margin confirms that our proposed Gated 1D representation is highly effective, capturing critical local biochemical features that conventional CNNs or graph convolutions may overlook.

Regarding the Davis dataset, LaPro-DTA yields the best ranking performance with a CI of 0.895 and the highest prediction stability with an $r_m^2$ of 0.693. Although AttentionDTA exhibits a marginal lead in MSE, our model demonstrates superior capability in correctly ranking drug-target pairs (indicated by CI) and maintaining predictive consistency (indicated by $r_m^2$). This suggests that LaPro-DTA is less biased towards the distribution mean and more robust in distinguishing relative affinity differences.
\begin{figure*}[!ht]
    \centering
    \includegraphics[width=\textwidth, trim=5pt 80pt 5pt 80pt, clip]{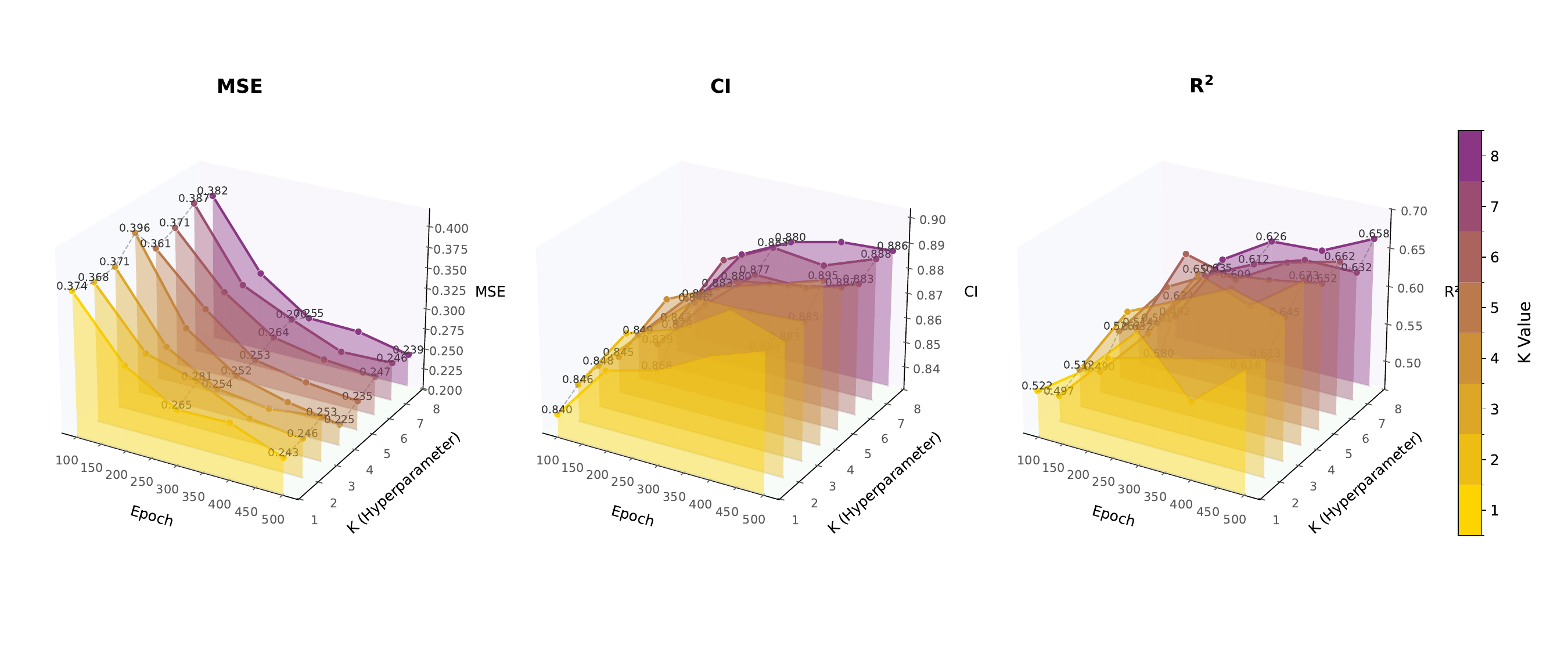}  % 左下右上 0.50 too wide
    \caption{Comparison of MSE, CI and $r_m^2$ of LaPro-DTA with different K on the Davis dataset.}
    \label{fig:K}
\end{figure*}
Furthermore, on the BindingDB dataset, which represents a much larger and more heterogeneous chemical space, LaPro-DTA consistently secures the top position in both MSE and CI. This evidence strongly supports the robustness of our framework, proving that the synergy of dual-view drug modeling and pattern-aware target pooling enables superior generalization across diverse molecular scaffolds and protein families.

Despite these challenges, LaPro-DTA demonstrates superior predictive capabilities. On the Davis dataset, our model achieves a substantial reduction in prediction error (MSE) compared to the strongest baseline, TransVAE-DTA, while establishing a new benchmark for ranking accuracy (CI). Although the predictive stability metric ($r_m^2$) remains comparable to the leading method, the distinct advantage in precision highlights the effectiveness of our feature fusion strategy.

On the larger KIBA dataset, the performance gap is even more pronounced. LaPro-DTA achieves comprehensive dominance, consistently outperforming previous state-of-the-art methods across all metrics. This indicates that the proposed dual-view mechanism and pattern-aware pooling not only enhance generalization in cold-start scenarios but also effectively refine feature granularity for standard interaction prediction tasks.

\subsection{Ablation Study and Parameter Sensitivity}
\noindent \textbf{Ablation Study.}
As detailed in the main text, Table \ref{tab:ablation_davis_full_rm2} presents the comprehensive ablation results under three cold-start settings. The results consistently indicate that removing any specific module—whether the latent dual-view mechanism, top-$k$ pooling, or cross-view fusion—leads to performance degradation across all evaluation metrics. This confirms that the complete framework is essential for achieving robust generalization when extrapolating to unseen biochemical entities.

To further verify the contribution of each module in standard evaluation scenarios, we conducted an ablation study on the Davis dataset using random splitting, as summarized in Table \ref{tab:ablation_random_split}. Comparing the baseline with the variant using only top-$k$ pooling reveals a sharp reduction in prediction error (MSE). This suggests that the top-$k$ strategy effectively filters out uninformative background noise in protein sequences, allowing the model to focus on high-confidence binding motifs even when training and testing data share similar distributions. Similarly, the introduction of the feature fusion module significantly boosts the concordance index, highlighting the critical role of attention-based interaction modeling in correctly ranking the relative affinities of drug-target pairs. Ultimately, the integration of the latent dual-view mechanism with these components yields the best overall performance. The dual-view approach complements local feature extraction by enforcing global structural consistency, which proves particularly beneficial for improving predictive stability. These results confirm that the proposed modules fundamentally enhance representation and reasoning capabilities beyond just cold-start generalization.

\noindent \textbf{Parameter Sensitivity.}
In addition to the cold-start experiments, we also conducted a sensitivity analysis on the key parameter $K$ under the random split setting on the Davis dataset. As illustrated in Figure~\ref{fig:K}, as $K$ increases from 1, the model performance exhibits a clear improvement in the early stage: MSE decreases substantially with increasing $K$, indicating reduced prediction error, while CI and $r_m^2$ consistently increase, reflecting simultaneous enhancements in ranking capability and goodness of fit. Notably, when $K > 1$, the model consistently outperforms the case of $K = 1$, demonstrating that the proposed top-$k$ Pooling effectively strengthens the representation of drug fragment-level features. As $K$ continues to increase, the variations of all three metrics gradually diminish, and the model performance converges to a stable regime without noticeable degradation.

\subsection{Experimental Analysis}

\noindent \textbf{Scatter Plots of Prediction Performance.}
Figure~\ref{fig:Scatter} illustrates the scatter plots of predicted versus observed binding affinities on the Davis and KIBA datasets under three cold-start settings. As shown, the predictions of LaPro-DTA are tightly clustered around the identity line across all scenarios, indicating strong consistency between predicted and observed affinities. Notably, the model exhibits particularly strong performance in the unseen drug setting on the KIBA dataset, where the scatter points are most concentrated. In contrast, a higher degree of dispersion is observed in the unseen target setting, suggesting that this scenario remains more challenging for accurate affinity prediction.

% \small
% \input{ijcai25.bbl}
\bibliographystyle{named}
\bibliography{ijcai25}

@article{chen2025local,
  title={{Local}--{Global} Structure-Aware Geometric Equivariant Graph Representation Learning for Predicting {Protein}--Ligand Binding Affinity},
  author={Chen, Shihong and Yi, Haicheng and You, Zhuhong and Shang, Xuequn and Huang, Yu-An and Wang, Lei and Wang, Zhen},
  journal={IEEE Transactions on Neural Networks and Learning Systems},
  year={2025},
  publisher={IEEE}
}

@article{li2023structure,
  title={Structure-aware graph attention diffusion network for protein--ligand binding affinity prediction},
  author={Li, Mei and Cao, Ye and Liu, Xiaoguang and Ji, Hua},
  journal={IEEE Transactions on Neural Networks and Learning Systems},
  year={2023},
  publisher={IEEE}
}

@article{kim2021bayesian,
  title={{Bayesian} neural network with pretrained protein embedding enhances prediction accuracy of drug-protein interaction},
  author={Kim, QHwan and Ko, Joon-Hyuk and Kim, Sunghoon and Park, Nojun and Jhe, Wonho},
  journal={Bioinformatics},
  volume={37},
  number={20},
  pages={3428--3435},
  year={2021},
  publisher={Oxford University Press}
}

@article{watanabe2021deep,
  title={Deep learning integration of molecular and interactome data for protein--compound interaction prediction},
  author={Watanabe, Narumi and Ohnuki, Yuuto and Sakakibara, Yasubumi},
  journal={Journal of Cheminformatics},
  volume={13},
  number={1},
  pages={36},
  year={2021},
  publisher={Springer}
}

@article{prasad2017research,
  title={Research and development spending to bring a single cancer drug to market and revenues after approval},
  author={Prasad, Vinay and Mailankody, Sham},
  journal={JAMA internal medicine},
  volume={177},
  number={11},
  pages={1569--1575},
  year={2017},
  publisher={American Medical Association}
}

@article{li2021co,
  title={{Co-VAE}: Drug-target binding affinity prediction by co-regularized variational autoencoders},
  author={Li, Tianjiao and Zhao, Xing-Ming and Li, Limin},
  journal={IEEE Transactions on Pattern Analysis and Machine Intelligence},
  volume={44},
  number={12},
  pages={8861--8873},
  year={2021},
  publisher={IEEE}
}

@article{ozturk2018deepdta,
  title={{DeepDTA}: deep drug--target binding affinity prediction},
  author={{\"O}zt{\"u}rk, Hakime and {\"O}zg{\"u}r, Arzucan and Ozkirimli, Elif},
  journal={Bioinformatics},
  volume={34},
  number={17},
  pages={i821--i829},
  year={2018},
  publisher={Oxford University Press}
}

@article{nguyen2021graphdta,
  title={{GraphDTA}: predicting drug--target binding affinity with graph neural networks},
  author={Nguyen, Thin and Le, Hang and Quinn, Thomas P and Nguyen, Tri and Le, Thuc Duy and Venkatesh, Svetha},
  journal={Bioinformatics},
  volume={37},
  number={8},
  pages={1140--1147},
  year={2021},
  publisher={Oxford University Press}
}

@inproceedings{zhao2019attentiondta,
  title={{AttentionDTA}: prediction of drug--target binding affinity using attention model},
  author={Zhao, Qichang and Xiao, Fen and Yang, Mengyun and Li, Yaohang and Wang, Jianxin},
  booktitle={2019 {IEEE} International Conference on Bioinformatics and Biomedicine ({BIBM})},
  pages={64--69},
  year={2019},
  organization={IEEE}
}

@article{zhou2024transvae,
  title={{TransVAE-DTA}: {Transformer} and variational autoencoder network for drug-target binding affinity prediction},
  author={Zhou, Changjian and Li, Zhongzheng and Song, Jia and Xiang, Wensheng},
  journal={Computer Methods and Programs in Biomedicine},
  volume={244},
  pages={108003},
  year={2024},
  publisher={Elsevier}
}

@article{zhang2024gdilateddta,
  title={{GDilatedDTA}: Graph dilation convolution strategy for drug target binding affinity prediction},
  author={Zhang, Longxin and Zeng, Wenliang and Chen, Jingsheng and Chen, Jianguo and Li, Keqin},
  journal={Biomedical Signal Processing and Control},
  volume={92},
  pages={106110},
  year={2024},
  publisher={Elsevier}
}

@article{huang2021moltrans,
  title={{MolTrans}: molecular interaction transformer for drug--target interaction prediction},
  author={Huang, Kexin and Xiao, Cao and Glass, Lucas M and Sun, Jimeng},
  journal={Bioinformatics},
  volume={37},
  number={6},
  pages={830--836},
  year={2021},
  publisher={Oxford University Press}
}

@article{ozturk2019widedta,
  title={{WideDTA}: prediction of drug-target binding affinity},
  author={{\"O}zt{\"u}rk, Hakime and Ozkirimli, Elif and {\"O}zg{\"u}r, Arzucan},
  journal={arXiv preprint arXiv:1902.04166},
  year={2019}
}

@inproceedings{dauphin2017language,
  title={Language modeling with gated convolutional networks},
  author={Dauphin, Yann N and Fan, Angela and Auli, Michael and Grangier, David},
  booktitle={International conference on machine learning},
  pages={933--941},
  year={2017},
  organization={PMLR}
}

@inproceedings{13,
  title={PAIR-VAE: Variational Pairwise Augmentation with Label Sharing for Generalizable Drug-Target Interaction Prediction},
  author={Qian, Yining and Zhang, Xuewen and Lu, An-Yang},
  booktitle={2025 IEEE International Conference on Bioinformatics and Biomedicine (BIBM)},
  pages={693--698},
  year={2025},
  organization={IEEE}
}

@article{nascimento2016multiple,
  title={A multiple kernel learning algorithm for drug-target interaction prediction},
  author={Nascimento, Andr{\'e} CA and Prud{\^e}ncio, Ricardo BC and Costa, Ivan G},
  journal={BMC Bioinformatics},
  volume={17},
  number={1},
  pages={46},
  year={2016},
  publisher={Springer}
}

@article{qian2023doublesg,
  title={{DoubleSG-DTA}: deep learning for drug discovery: case study on the non-small cell lung cancer with {EGFR} {T} 790 {M} mutation},
  author={Qian, Yongtao and Ni, Wanxing and Xianyu, Xingxing and Tao, Liang and Wang, Qin},
  journal={Pharmaceutics},
  volume={15},
  number={2},
  pages={675},
  year={2023},
  publisher={MDPI}
}

@article{davis2011comprehensive,
  title={Comprehensive analysis of kinase inhibitor selectivity},
  author={Davis, Mindy I and Hunt, Jeremy P and Herrgard, Sanna and Ciceri, Pietro and Wodicka, Lisa M and Pallares, Gabriel and Hocker, Michael and Treiber, Daniel K and Zarrinkar, Patrick P},
  journal={Nature Biotechnology},
  volume={29},
  number={11},
  pages={1046--1051},
  year={2011},
  publisher={Nature Publishing Group US New York}
}

@article{tang2014making,
  title={Making sense of large-scale kinase inhibitor bioactivity data sets: a comparative and integrative analysis},
  author={Tang, Jing and Szwajda, Agnieszka and Shakyawar, Sushil and Xu, Tao and Hintsanen, Petteri and Wennerberg, Krister and Aittokallio, Tero},
  journal={Journal of chemical information and modeling},
  volume={54},
  number={3},
  pages={735--743},
  year={2014},
  publisher={ACS Publications}
}

@article{liu2007bindingdb,
  title={{BindingDB}: a web-accessible database of experimentally determined protein--ligand binding affinities},
  author={Liu, Tiqing and Lin, Yuhmei and Wen, Xin and Jorissen, Robert N and Gilson, Michael K},
  journal={Nucleic acids research},
  volume={35},
  number={suppl\_1},
  pages={D198--D201},
  year={2007},
  publisher={Oxford University Press}
}

@article{yang2024interaction,
  title={Interaction-based inductive bias in graph neural networks: enhancing protein-ligand binding affinity predictions from {3D} structures},
  author={Yang, Ziduo and Zhong, Weihe and Lv, Qiujie and Dong, Tiejun and Chen, Guanxing and Chen, Calvin Yu-Chian},
  journal={IEEE Transactions on Pattern Analysis and Machine Intelligence},
  volume={46},
  number={12},
  pages={8191--8208},
  year={2024},
  publisher={IEEE}
}

@inproceedings{li2020monn,
  title={{MONN}: A multi-objective neural network for predicting pairwise non-covalent interactions and binding affinities between compounds and proteins},
  author={Li, Shuya and Wan, Fangping and Shu, Hantao and Jiang, Tao and Zhao, Dan and Zeng, Jianyang},
  booktitle={International Conference on Research in Computational Molecular Biology},
  pages={259--260},
  year={2020},
  organization={Springer}
}

@article{zhao2020gansdta,
  title={{GANsDTA}: Predicting drug-target binding affinity using {GANs}},
  author={Zhao, Lingling and Wang, Junjie and Pang, Long and Liu, Yang and Zhang, Jun},
  journal={Frontiers in genetics},
  volume={10},
  pages={1243},
  year={2020},
  publisher={Frontiers Media SA}
}

@article{fang2023colddta,
  title={{ColdDTA}: utilizing data augmentation and attention-based feature fusion for drug-target binding affinity prediction},
  author={Fang, Kejie and Zhang, Yiming and Du, Shiyu and He, Jian},
  journal={Computers in Biology and Medicine},
  volume={164},
  pages={107372},
  year={2023},
  publisher={Elsevier}
}

@article{ain2015machine,
  title={Machine-learning scoring functions to improve structure-based binding affinity prediction and virtual screening},
  author={Ain, Qurrat Ul and Aleksandrova, Antoniya and Roessler, Florian D and Ballester, Pedro J},
  journal={Wiley Interdisciplinary Reviews: Computational Molecular Science},
  volume={5},
  number={6},
  pages={405--424},
  year={2015},
  publisher={Wiley Online Library}
}

@article{he2017simboost,
  title={{SimBoost}: a read-across approach for predicting drug--target binding affinities using gradient boosting machines},
  author={He, Tong and Heidemeyer, Marten and Ban, Fuqiang and Cherkasov, Artem and Ester, Martin},
  journal={Journal of cheminformatics},
  volume={9},
  number={1},
  pages={24},
  year={2017},
  publisher={Springer}
}

@article{pratim2009two,
  title={On two novel parameters for validation of predictive {QSAR} models},
  author={Pratim Roy, Partha and Paul, Somnath and Mitra, Indrani and Roy, Kunal},
  journal={Molecules},
  volume={14},
  number={5},
  pages={1660--1701},
  year={2009},
  publisher={Molecular Diversity Preservation International}
}

@article{bi2023hisif,
  title={{HiSIF-DTA}: a hierarchical semantic information fusion framework for drug-target affinity prediction},
  author={Bi, Xiangpeng and Zhang, Shugang and Ma, Wenjian and Jiang, Huasen and Wei, Zhiqiang},
  journal={IEEE Journal of Biomedical and Health Informatics},
  year={2023},
  publisher={IEEE}
}

@inproceedings{selvaraju2017grad,
  title={{Grad-CAM}: Visual explanations from deep networks via gradient-based localization},
  author={Selvaraju, Ramprasaath R and Cogswell, Michael and Das, Abhishek and Vedantam, Ramakrishna and Parikh, Devi and Batra, Dhruv},
  booktitle={Proceedings of the {IEEE} international conference on computer vision},
  pages={618--626},
  year={2017}
}

@article{humphrey1996vmd,
  title={{VMD}: visual molecular dynamics},
  author={Humphrey, William and Dalke, Andrew and Schulten, Klaus},
  journal={Journal of molecular graphics},
  volume={14},
  number={1},
  pages={33--38},
  year={1996},
  publisher={Elsevier}
}

@inproceedings{zeiler2010deconvolutional,
  title={Deconvolutional networks},
  author={Zeiler, Matthew D and Krishnan, Dilip and Taylor, Graham W and Fergus, Rob},
  booktitle={2010 {IEEE} Computer Society Conference on computer vision and pattern recognition},
  pages={2528--2535},
  year={2010},
  organization={IEEE}
}

@article{weininger1988smiles,
  title={{SMILES}, a chemical language and information system. 1. Introduction to methodology and encoding rules},
  author={Weininger, David},
  journal={Journal of chemical information and computer sciences},
  volume={28},
  number={1},
  pages={31--36},
  year={1988},
  publisher={ACS Publications}
}

@article{huang2025gflearn,
  title={{GFLearn}: Generalized Feature Learning for Drug-Target Binding Affinity Prediction},
  author={Huang, Zibo and Weng, Xinrui and Ou-Yang, Le},
  journal={IEEE Journal of Biomedical and Health Informatics},
  year={2025},
  publisher={IEEE}
}

@article{choppara2025q,
  title={{Q-BAFNet}: A Hybrid Quantum Classical Approach for Drug-Target Binding Affinity Prediction},
  author={Choppara, Prashanth and Lokesh, Bommareddy},
  journal={IEEE Transactions on Computational Biology and Bioinformatics},
  year={2025},
  publisher={IEEE}
}

@inproceedings{kalchbrenner2014convolutional,
  title={A Convolutional Neural Network for Modelling Sentences},
  author={Kalchbrenner, Nal and Grefenstette, Edward and Blunsom, Phil},
  booktitle={Proceedings of the 52nd Annual Meeting of the Association for Computational Linguistics ({ACL})},
  pages={655--665},
  year={2014}
}

@article{bemis1996properties,
  title={The properties of known drugs. 1. {Molecular} frameworks},
  author={Bemis, Guy W and Murcko, Mark A},
  journal={Journal of Medicinal Chemistry},
  volume={39},
  number={15},
  pages={2887--2893},
  year={1996},
  publisher={ACS Publications}
}

@article{Bishop1995,
  title={Training with noise is equivalent to {Tikhonov} regularization},
  author={Bishop, Chris M},
  journal={Neural computation},
  volume={7},
  number={1},
  pages={108--116},
  year={1995},
  publisher={MIT Press}
}

@article{shah2025deepdtagen,
  title={{DeepDTAGen}: a multitask deep learning framework for drug-target affinity prediction and target-aware drugs generation},
  author={Shah, Pir Masoom and Zhu, Huimin and Lu, Zhangli and Wang, Kaili and Tang, Jing and Li, Min},
  journal={Nature Communications},
  volume={16},
  number={1},
  pages={5021},
  year={2025},
  publisher={Nature Publishing Group UK London}
}

@article{abbasi2020deepcda,
  title={{DeepCDA}: deep cross-domain compound--protein affinity prediction through LSTM and convolutional neural networks},
  author={Abbasi, Karim and Razzaghi, Parvin and Poso, Antti and Amanlou, Massoud and Ghasemi, Jahan B and Masoudi-Nejad, Ali},
  journal={Bioinformatics},
  volume={36},
  number={17},
  pages={4633--4642},
  year={2020},
  publisher={Oxford University Press}
}

@article{he2025dual,
  title={Dual modality feature fused neural network integrating binding site information for drug target affinity prediction},
  author={He, Haohuai and Chen, Guanxing and Tang, Zhenchao and Chen, Calvin Yu-Chian},
  journal={NPJ Digital Medicine},
  volume={8},
  number={1},
  pages={67},
  year={2025},
  publisher={Nature Publishing Group UK London}
}

@article{vandermaaten2008visualizing,
  title={Visualizing data using t-SNE},
  author={van der Maaten, Laurens and Hinton, Geoffrey},
  journal={Journal of Machine Learning Research},
  volume={9},
  number={11},
  pages={2579--2605},
  year={2008}
}

\end{document}